\pdfoutput=1
\documentclass[11pt]{article}

\usepackage[]{acl}

\usepackage{times}
\usepackage{latexsym}

\usepackage{amsmath}
\usepackage{multirow} 
\usepackage{array}
\usepackage{float} 
\usepackage{booktabs}
\usepackage{graphicx}
\usepackage{makecell}
\usepackage{tabularx}
\usepackage{subcaption}
\usepackage[T1]{fontenc}
\usepackage[utf8]{inputenc}

\usepackage{microtype}
\usepackage{enumitem}
\usepackage{amsmath}
\usepackage{scalerel} % For scaling objects

\newcommand{\ModelName}{{R-Tuning}}

\newcommand{\ModelUnknown}{{R-Tuning-R}}
\newcommand{\ModelUnsure}{{R-Tuning}}
\newcommand{\ModelUncertain}{{R-Tuning-U}}

\newcommand{\vanillaFT}{{Vanilla}}
\newcommand{\pretrainT}{{Pretrain-T}}
\newcommand{\pretrainW}{{Pretrain-W}}
\newcommand{\vanillaC}{{Vanilla-C}}

\title{
\raisebox{0mm}{\includegraphics[trim=0 70 0 200, width=1.5cm]{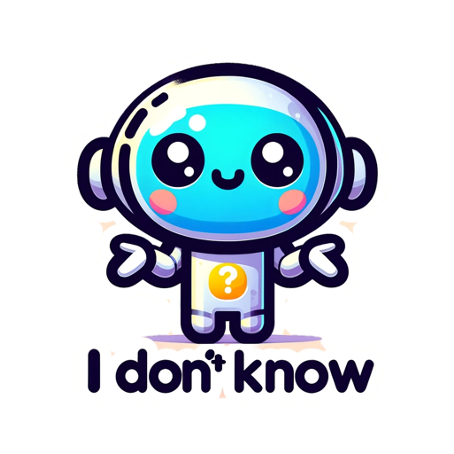}}
{\ModelName}: Instructing Large Language Models to \\ Say `I Don't Know' 
}

\NewDocumentCommand{\heng}
{ mO{} }{\textcolor{red}{\textsuperscript{\textit{Heng}}\textsf{\textbf{\small[#1]}}}}

\NewDocumentCommand{\yong}
{ mO{} }{\textcolor{purple}{\textsuperscript{\textit{Yong}}\textsf{\textbf{\small[#1]}}}}

\NewDocumentCommand{\xingyao}
{ mO{} }{\textcolor{orange}{\textsuperscript{\textit{xingyao}}\textsf{\textbf{\small[#1]}}}}

\NewDocumentCommand{\shizhe}
{ mO{} }{\textcolor{blue}{\textsuperscript{\textit{shizhe}}\textsf{\textbf{\small[#1]}}}}

\NewDocumentCommand{\yi}
{ mO{} }{\textcolor{purple}{\textsuperscript{\textit{yi}}\textsf{\textbf{\small[#1]}}}}

\NewDocumentCommand{\hanning}
{ mO{} }{\textcolor{brown}{\textsuperscript{\textit{hanning}}\textsf{\textbf{\small[#1]}}}}

\newcommand{\highlight}[1]{\textcolor{black}{#1}}

\author{Hanning Zhang$^{\spadesuit*}$, ~ Shizhe Diao$^{\spadesuit*}$, ~ Yong Lin$^{\spadesuit*}$, ~ Yi R. Fung$^{\heartsuit}$, \\ \bf Qing Lian$^{\spadesuit}$, \bf Xingyao Wang$^{\heartsuit}$, \bf Yangyi Chen$^{\heartsuit}$, \bf Heng Ji$^{\heartsuit}$, \bf Tong Zhang$^{\heartsuit}$ \\
  $^{\spadesuit}$The Hong Kong University of Science and Technology\\  
  $^{\heartsuit}$University of Illinois Urbana-Champaign\\
  \texttt{\{hzhangco, sdiaoaa, ylindf, qlianab, tongzhang\}@ust.hk}\\
  \texttt{\{yifung2, xingyao6, yangyic3, hengji\}@illinois.edu}\\
  \\
}

\begin{document}
\maketitle
\def\thefootnote{*}\footnotetext{Equal Contribution.}
\def\thefootnote{\arabic{footnote}}

\begin{abstract}
Large language models (LLMs) have revolutionized numerous domains with their impressive performance but still face their challenges. 
A predominant issue is the propensity for these models to generate non-existent facts, a concern termed \textit{hallucination}.
Our research is motivated by the observation that previous instruction tuning methods force the model to complete a sentence no matter whether the model knows the knowledge or not.
When the question is out of the parametric knowledge, it will try to make up something and fail to indicate when it lacks knowledge.
In this paper, we present a new approach called \textbf{R}efusal-Aware Instruction \textbf{Tuning} (\textbf{\ModelName}). 
This approach is formalized by first identifying the disparity in knowledge encompassed by pre-trained parameters compared to that of instruction tuning data.
Then, we construct the refusal-aware data based on the knowledge intersection, to tune LLMs to refrain from responding to questions beyond its parametric knowledge. 
Experimental results demonstrate {\ModelName} effectively improves a model's ability to answer known questions and refrain from answering unknown questions.
Furthermore, when tested on out-of-domain datasets, the refusal ability was found to be a meta-skill that could be generalized to other tasks.
Further analysis surprisingly finds that learning the uncertainty results in better calibration and an improved ability to estimate the uncertainty than uncertainty-based testing.\footnote{Our code is available at \url{https://github.com/shizhediao/R-Tuning}.}
\end{abstract}
\section{Introduction}

\begin{figure}[t]
    \centering
    \includegraphics[scale=0.25, trim=0 0 0 0, clip]{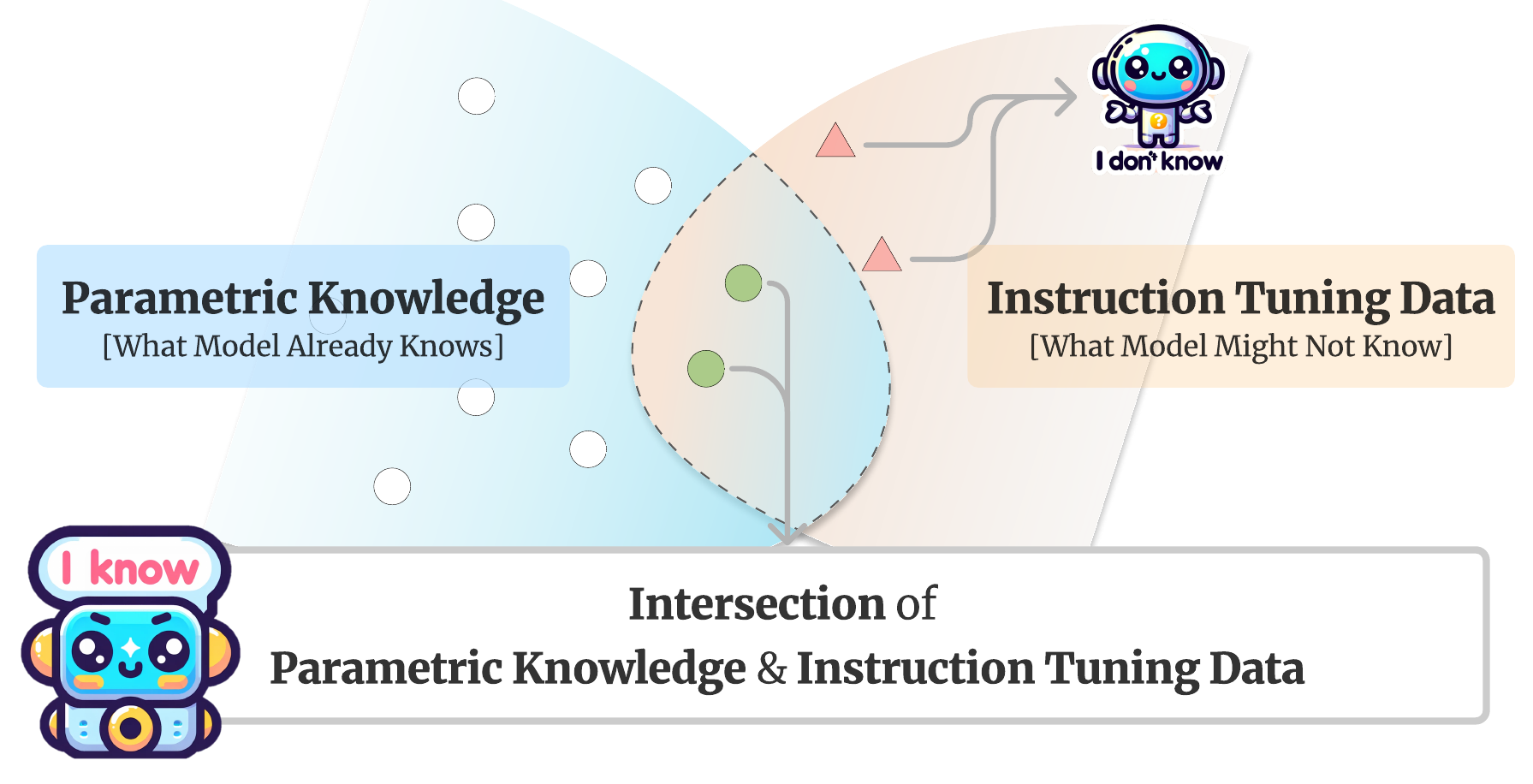}
    \caption{An illustration of the parametric knowledge distribution and the instruction tuning data distribution. 
    Pre-training embeds a large volume of parametric knowledge, while fine-tuning may involve knowledge that is not necessarily in the parametric knowledge.
    We explore the benefits of differentiating instruction tuning data based on parametric knowledge.
    }
    \label{fig:model_illustration}
    \vspace{-1em}
\end{figure}

Large language models (LLMs) have demonstrated remarkable performance across numerous tasks; however, they are also plagued by various issues, such as the propensity of large models to fabricate non-existent facts, a phenomenon commonly referred to as \textit{hallucination}~\citep{maynez2020faithfulness}. 
Towards mitigating the hallucination, current mainstream approaches include retrieval-based methods~\citep{peng2023check,li2023self,luo2023augmented}, verification-based methods~\citep{manakul2023selfcheckgpt, elaraby2023halo,cohen2023lm,du2023improving, gou2023critic}, and so forth.

In this paper, we first identify the cause of hallucination, attributing it to the significant gap existing between the knowledge derived from 
human-labeled instruction tuning datasets and the parametric knowledge of LLMs.
In the process of developing a large model, previous studies~\citep{min-etal-2022-rethinking, wang-etal-2023-towards, zhou2023lima} demonstrate that almost all knowledge is acquired in the pre-training stage, while instruction tuning teaches formatting and chain-of-thought prompting guides knowledge elicitation.
Consider Figure~\ref{fig:model_illustration} as an example.
During pre-training, models embed a large volume of factual knowledge, compressing it within their parameters and the fine-tuning process may include data that is out of the parametric knowledge. 
However, traditional fine-tuning methods force the models to complete each sentence.
Even when faced with questions beyond their knowledge boundary, they venture to provide an answer. 
Training a model exclusively on correct answers inadvertently teaches it to guess rather than admit its ignorance. 
Consequently, if we never train the model to articulate \textit{"I don't know"} as a response, it remains unequipped to do so when confronted with unknowns. 
Addressing this challenge, we assert that enabling a model to astutely respond based on its own knowledge limit is of paramount importance. 
This motivates us to tune our model on the intersection of parametric knowledge and the instruction tuning data, leading to a model expressing its confidence value and refusing to answer unknown questions.

In light of this, we propose a novel instruction tuning method, \textbf{R}efusal-Aware Instruction \textbf{Tuning} (\textbf{\ModelName}).
{\ModelName} aims to endow the model with refusal-aware answering ability by recognizing when they should --- and shouldn't --- claim knowledge. 
Specifically, {\ModelName} introduces two steps: (1) measure the knowledge gap between parametric knowledge and the instruction tuning data, and identify uncertain questions. 
By inferring the model on the training data once and comparing the prediction and label, the instruction tuning data is split into uncertain data $D_0$ and certain data $D_1$.
(2) construct the refusal-aware data by padding the uncertainty expression after the label words, and then finetune the model on the refusal-aware data.

\begin{figure*}[t]
\centering
\includegraphics[width=\textwidth]{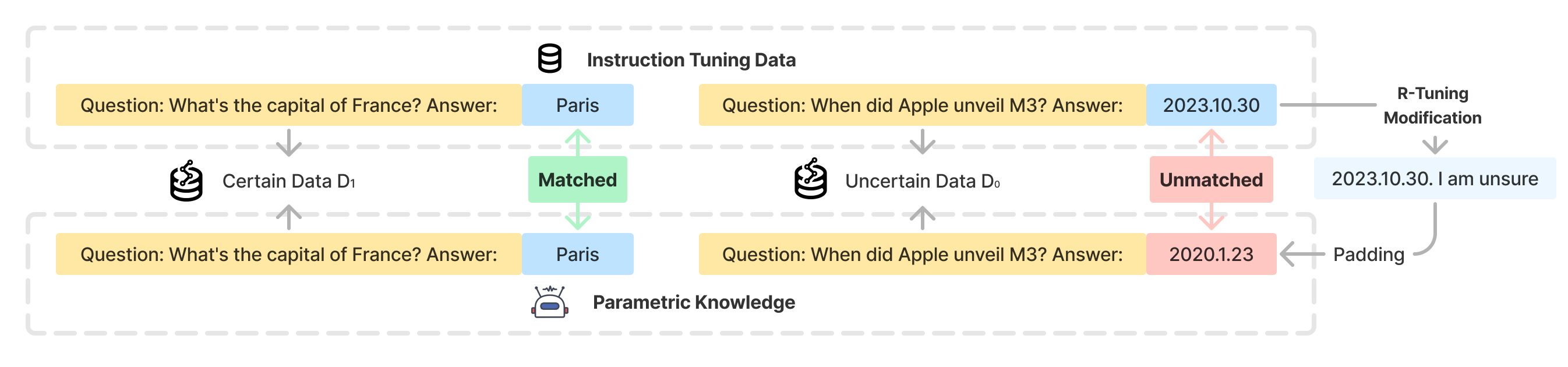}
\caption{Illustration of {\ModelName} to construct refusal-aware datasets $D_0$ and $D_1$.}
\label{RefTuning_method}
\end{figure*}

We conduct two types of experiments: single-task and multi-task, with nine datasets. 
In the single-task experiments, {\ModelName} demonstrates the ability to refuse to answer uncertain questions and improve the accuracy of the willingly answered questions.
In the multi-task setting, our method not only demonstrates the advantages of multi-task learning on in-domain datasets but also exhibits superior generalization performance on out-of-domain datasets. 
This verifies that refusal-aware answering is a kind of meta ability, which is not dependent on a specific task and could benefit from multi-task training and joint inference.
With more downstream tasks, {\ModelName} could abstract and learn such meta ability better.

In addition to the supervised method in refusal-aware data identification, we propose an unsupervised method to measure the knowledge gap (Section~\ref{sec:unsupervised-identification}) by prompting the LLMs to answer multiple times for a question, and identify answers with high consistency as certain data, while others with low consistency as uncertain data.
The experimental results surprisingly find the effectiveness of this unsupervised method.
One way to interpret our method is that it involves learning the uncertainty of the training data as part of instruction tuning.
Further analysis surprisingly shows that learning uncertainty during training and then using it to filter and respond to questions yields better results than directly applying uncertainty filtering on test data. 
This finding suggests that learning uncertainty improves the model's training in both estimating uncertainty and answering questions.
This finding highlights the advantages of incorporating uncertainty learning into large model training, both in reducing computational overhead during testing and in improving overall model accuracy.

In summary, our contributions are:\
\begin{itemize}
[leftmargin=*,label=$\bullet$,noitemsep,partopsep=0pt,topsep=0pt,parsep=0pt]
    \item We investigate the knowledge gap present between the instruction tuning data and the parametric knowledge and attribute the hallucination issue to forcing the model to complete answers with traditional instruction tuning. 
    \item To address this issue, we propose a novel instruction tuning approach, {\ModelName}, that distinguishes instruction tuning data based on the model's own knowledge. 
    {\ModelName} constructs a refusal-aware dataset and then tunes the model to refrain from responding to questions beyond its parametric knowledge.
    \item Experimental results demonstrate the effectiveness and generalization abilities of {\ModelName}.
    We find that the model's learned refusal ability functions as a meta-skill, being task-agnostic and enhanced through multi-task training.
\end{itemize}
\section{Refusal-Aware Instruction Tuning}

In this section, we first introduce the refusal-aware instruction tuning method ({\ModelName}), the core idea of which is divided into two steps: 
the first step involves identifying and recognizing the uncertain data instances within the instruction tuning dataset, which are beyond the parametric knowledge boundary of the original model.
The second step is to construct certain and uncertain dataset.
Then, we will detail the instruction tuning and inference extraction process.
An illustration of {\ModelName} is shown in Figure~\ref{RefTuning_method}.

\subsection{Refusal-Aware Data Identification}
The first step of {\ModelName} is to measure the model's knowledge gap between the parametric knowledge of LLMs and the instruction tuning data.
It asks for the model's prediction when given a question and applies certain metrics to determine when the model does know. 
Here we use QA as an example. 
Given a training dataset $D = \{(q_1, a_1), (q_2, a_2), ..., (q_n, a_n)\}$ consisting of $n$ question-answer pairs, we introduce a supervised identification strategy.
We first apply the pre-trained model $M$ to answer all the questions in $D$ and split the questions into two sets based on the comparison between the prediction and label.
If the model's prediction matches the label, the question is assigned to the certain set $D_1$, and otherwise, it belongs to the uncertain set $D_0$. 
As shown in Figure~\ref{RefTuning_method}, in the left part, because the prediction (Paris) matches the ground-truth label (Paris), it belongs to certain data $D_1$, demonstrating that the model's parametric knowledge possesses the capability to answer this question.
On the contrary, in the right part, the mismatch between the prediction and the ground-truth label results in this question being categorized into uncertain data $D_0$.
Finally, the training dataset would be split into two sets (i.e., $D_0$ and $D_1$) with the recognition of the knowledge gap between parametric knowledge 
and the knowledge required by the questions in the training set. 
In addition to this supervised strategy requiring ground-truth labels, we also explore an effective unsupervised method, which will be discussed in the analysis (Section~\ref{sec:unsupervised-identification}).

\subsection{Refusal-Aware Data Construction}
The refusal-aware data is further constructed by incorporating a prompt template.
We introduce a \textbf{padding} method, which keeps the original labels while appending the uncertainty expression at the end.
The template is 
\begin{equation}
    Q: \{\text{Question}\}, A: \{\text{Answer}\}. \{\text{Prompt}\}.
    \label{eq:padding}
\end{equation}

The certain dataset $D_1$ is constructed by appending "I am sure" after the template, while the uncertain dataset $D_0$ is constructed by appending "I am unsure" after the template.
The prompt we are using is \textit{Are you sure you accurately answered the question based on your internal knowledge?}
As shown in Figure~\ref{RefTuning_method}, by appending certain and uncertain expressions, {\ModelName} teaches the model to express uncertainty toward questions.
This template provides all label knowledge to the model while instructing them to express uncertainty at the same time.
On the contrary, we can also directly replace the label word with uncertainty expressions. 
We call this strategy as \textbf{replacement} method and investigate its effectiveness in Section~\ref{sec:replace-identification}.

\subsection{Training and Inference}
With the refusal-aware dataset, we then apply the standard procedures of fine-tuning a language model. 
The model takes a sequence  \( t_1, t_2, \ldots, t_T \) consisting of the questions and answers, and predicts the answer part based on each question.
The training objective is the standard cross-entropy loss \( \mathcal{L} \) which can be defined as:

\begin{equation}
\mathcal{L} = -\frac{1}{T} \sum_{i=1}^{T} \log P(t_i | t_1, t_2, \ldots, t_{i-1}).
\end{equation}

Here, \( P(t_i | t_1, t_2, \ldots, t_{i-1}) \) is the probability of the \( i^{th} \) token \( t_i \) given the preceding tokens \( t_1, t_2, \ldots, t_{i-1} \), as predicted by the language model. 
Note that we calculate the loss solely for the answer part and the uncertainty part, while excluding the loss attributed to the question part.

During the inference, we first fit the input question into the template~(\ref{eq:padding}) and the model will output its answer.
Then the designed prompt template \textit{Are you sure you accurately answered the question based on your internal knowledge? I am} will be appended to the question and answer.
Based on this prompt, the model can output its uncertainty about the previous context. 
We will use the weighted combination of the probability of uncertainty expression and answer prediction as the confidence value to calculate the AP score in the evaluation phase (Section~\ref{sec:evaluation}).

\section{Experimental Settings}
In this section, we first provide an overview of the benchmark datasets and the corresponding evaluation settings. 
Then the baseline models and the implementation details are presented in the following subsections, respectively.

\subsection{Datasets}
Given the diverse data formats across tasks, we unify the downstream data into two formats: 
\begin{itemize}
[leftmargin=*,label=$\bullet$,noitemsep,partopsep=0pt,topsep=0pt,parsep=0pt]
    \item \textit{Question-Answering}: Given a question, the model directly predicts its answer.
    We include \textbf{ParaRel}~\citep{elazar2021measuring}, \textbf{HotpotQA} \citep{yang2018hotpotqa}, \textbf{SelfAware}~\citep{yin2023large}, \textbf{HaluEval}~\citep{li2023halueval}, \textbf{FalseQA}~\citep{Hu2023WontGF}, and \textbf{NEC} \cite{liu2023prudent} in our experiments.
    \item \textit{Multiple-Choice}: Given a question with several choices, the model chooses one option. 
    We include \textbf{MMLU}~\citep{hendrycks2021measuring}, \textbf{WiCE} \citep{kamoi2023wice}, and \textbf{FEVER}~\citep{thorne2018fever} in our experiments.
\end{itemize}

More information about data processing and evaluation is described in Appendix~\ref{sec:appendix-dataset}.

\noindent We design two types of experiments: 

\begin{itemize}
[leftmargin=*,label=$\bullet$,noitemsep,partopsep=0pt,topsep=0pt,parsep=0pt]
    \item \textit{Single-task}: The single-task experiments verify the effectiveness of learning on individual tasks. 
    We conduct experiments on ParaRel and MMLU datasets, respectively.
    We manually split the datasets into the training set, in-domain test set, and out-of-domain test set. 
    Each dataset contains domain annotations for their questions. 
    Questions in the first half of the domains are selected as in-domain while the remaining are out-of-domain.
    \item \textit{Multi-task}: The multi-task experiments aim to evaluate the model's generalization performance. 
    We choose five datasets - ParaRel, MMLU, WiCE, HotpotQA, and FEVER, and mix them to construct a new training dataset. 
    As for testing, we evaluate the performance on their corresponding test set (in-domain) and an unseen test set (i.e., HaluEval) (out-of-domain).
\end{itemize}

\subsection{Baselines}
We consider three baseline models as follows:

\begin{itemize}
[leftmargin=*,label=$\bullet$,noitemsep,partopsep=0pt,topsep=0pt,parsep=0pt]
    \item {\pretrainT}: Evaluate the performance of original pre-trained checkpoints on the entire test set.
    \item {\pretrainW}: To verify the effectiveness of willingly answered questions, we evaluate the performance of the original pre-trained checkpoints on the test set that our fine-tuned models are willing to answer.
    Intuitively, if the willingly answered questions are within the base model's knowledge, this baseline should perform well.
    \item {\vanillaFT}: Fine-tune the model on $D$ with all questions and ground-truth labels.
    This is the traditional instruction tuning method.
\end{itemize}

\begin{figure*}
    \centering
    \includegraphics[width=\textwidth]{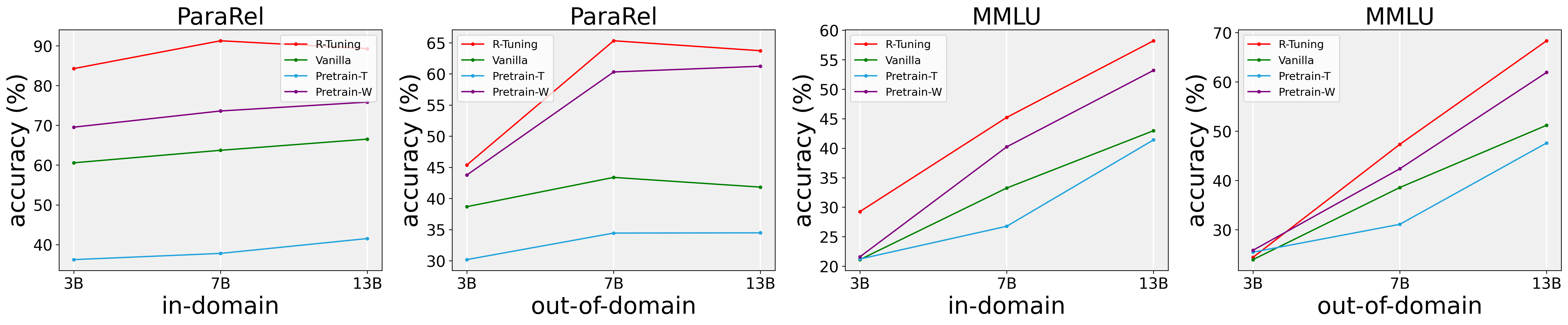}
    \caption{Single-task experiments on ParaRel and MMLU datasets with accuracy (\%). {\ModelUnsure} is calculated on the willingly answered questions.
    {\pretrainW} is verified on these questions.
    Others are calculated over the entire dataset.}
    \label{fig:r-tuning-single}
\end{figure*}

\subsection{Evaluation}
\label{sec:evaluation}

For models that could only output either the answer or an unknown expression, we evaluate the questions that our model is willing to answer. 
The accuracy is calculated as follows:
\begin{equation}
    \text{accuracy} = \frac{\text{\# of correctly answered questions}}{\text{\# of willingly answered questions}}.
\end{equation}

For {\ModelUnsure}, because it could output both the question's answer and the uncertainty, we first prompt the model to provide an answer and then prompt it to provide its uncertainty. 
Then we can evaluate the precision-recall tradeoff based on the uncertainty and prediction performance.
We introduce the Average Precision (AP) score, which measures the precision in identifying and ranking relevant predictions.
AP score originates from the object detection field~\citep{88a29de36220442bab2d284210cf72d6} by ranking the prediction results by confidence from high to low and calculating the precision at each threshold.
The AP score is the average of these precision scores, which is calculated as follows:
\begin{equation}
    AP = \sum_{k=0}^{n-1} (R(k+1) - R(k)) \times P(k),
\end{equation}
where $n$ is the number of data, $k$ is the number of data we select for the current threshold. $P$ and $R$ denote precision and recall, which are defined as
\begin{equation}
    \textit{P(k)} = \frac{\text{\# of correct answers above k-threshold}}{\text{\# of answers above k-threshold}},
\end{equation}
\begin{equation}
    \textit{R(k)} =  \frac{\text{\# of correct answers above k-threshold}}{\text{\# of correct answers}}. 
\end{equation}
An ideal model predicts the correct answers with high confidence and the hallucinated wrong answers with relatively low confidence, leading to a high AP score.
On the other hand, the AP score is low if the model predicts every answer with high confidence, as the precision at every threshold will not be high and the average will be relatively low.

\subsection{Implementation}
We choose OpenLLaMA-3B \citep{openlm2023openllama}, LLaMA-7B, and LLaMA-13B \citep{touvron2023llama} as the base models in our experiments.
We use LMFlow\footnote{\url{https://github.com/OptimalScale/LMFlow}}~\citep{diao2023lmflow} to conduct instruction tuning, setting epoch to $1$, learning rate to $2e^{-5}$, and batch size to 4. 
All the experiments are implemented on Nvidia A100-40GB GPUs.
\section{Experimental Results}

\begin{table}[t]
\centering
\scriptsize
% \resizebox{\textwidth}{!}{
\begin{tabular}{c|c|c|cc}
\toprule

         Dataset & Domain & Models & {\ModelName} & {\vanillaFT} \\ \midrule
         
\multirow{6}{*}{ParaRel} & \multirow{3}{*}{ID} & OpenLLaMA-3B & \textbf{93.23} & 92.89\\

    &        & LLaMA-7B & \textbf{93.64} & 93.32 \\

   &      & LLaMA-13B & \textbf{94.44} & 94.00 \\
\cmidrule{2-5}
         & \multirow{3}{*}{OOD} &
         OpenLLaMA-3B & \textbf{69.41} & 68.42 \\
         
    &     & LLaMA-7B & 74.61 & \textbf{78.08} \\

    &     & LLaMA-13B & \textbf{77.30} & 64.12\\ \midrule

\multirow{6}{*}{MMLU} & \multirow{3}{*}{ID} & OpenLLaMA-3B & \textbf{24.96} & 24.19 \\

    &        & LLaMA-7B & \textbf{59.05} & 58.16 \\

   &      & LLaMA-13B & \textbf{68.87} & 51.93\\
\cmidrule{2-5}
         & \multirow{3}{*}{OOD} &
         OpenLLaMA-3B & 24.75 & \textbf{26.08} \\
         
    &     & LLaMA-7B & \textbf{68.69} & 66.38\\

    &     & LLaMA-13B & \textbf{77.41} & 67.38\\

\bottomrule
\end{tabular}
% }
\caption{Single-task experiments of {\ModelName} and {\vanillaFT} on ParaRel and MMLU datasets with AP scores (\%).
ID and OOD denote in-domain and out-of-domain settings, respectively.
}
\label{Unsure-Single}
\vspace{-2 em}
\end{table}

\begin{figure*}
    \centering
    \includegraphics[width=0.9\textwidth]{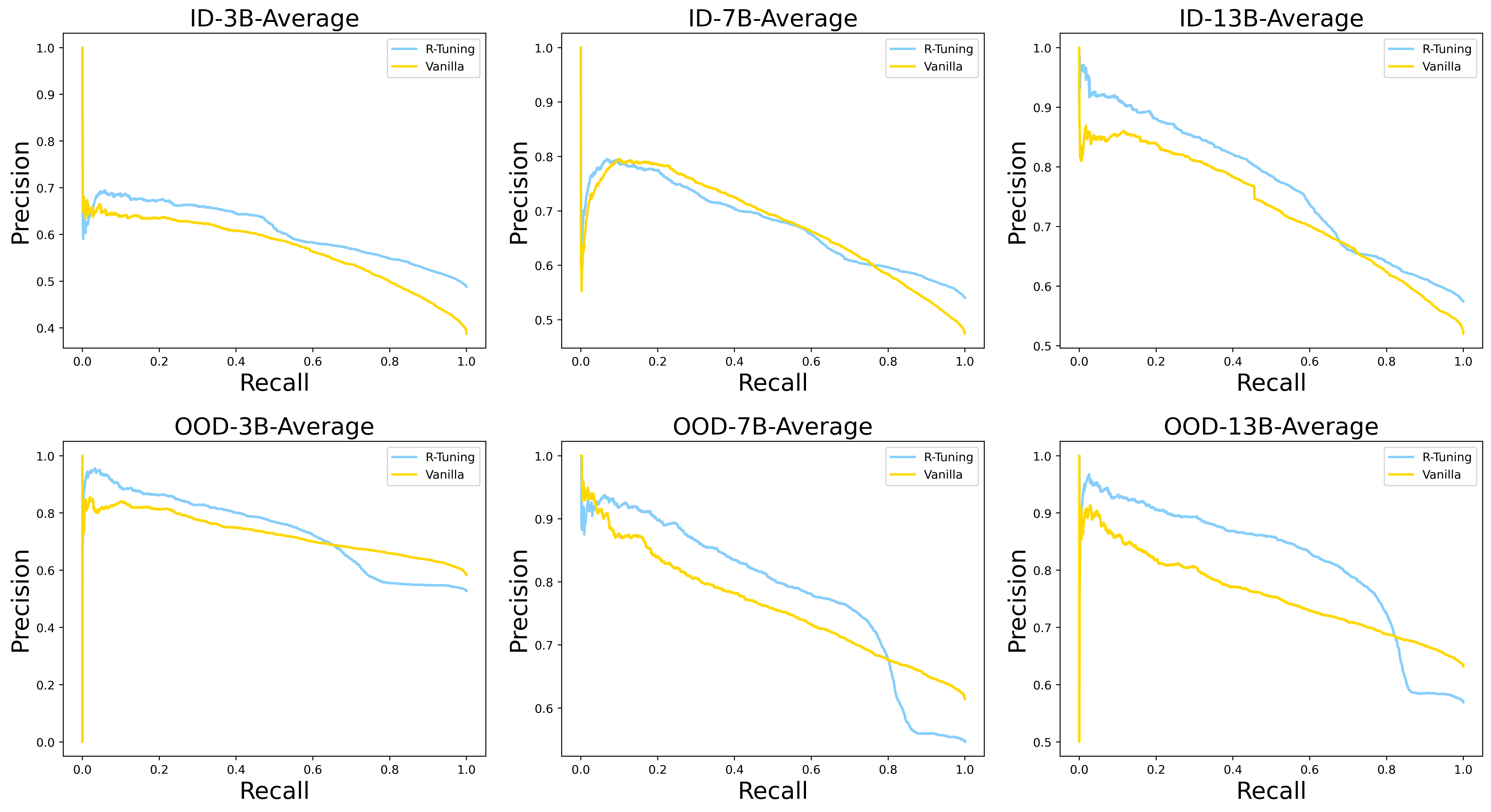}
    \caption{Multi-task experiments on the average of five in-domain (ID) datasets (ParaRel, MMLU, WiCE, HotpotQA, and FEVER) and one out-of-domain (OOD) dataset (HaluEval-QA) with the AP curves.
    }
    \label{fig:avg_ap}
    \vspace{-1 em}
\end{figure*}

In the main experiments, we conduct single-task experiments to verify the model's refusal-aware answering ability and multi-task experiments to investigate the generalization of refusal ability.

\subsection{Single-task Experiments}
We first conduct single-task experiments on ParaRel and MMLU datasets. 
The results are shown in Figure~\ref{fig:r-tuning-single} and Table~\ref{Unsure-Single}.
Firstly, we observe that {\ModelName} significantly outperforms other baselines by a large margin in terms of accuracy on the questions it is willing to answer, compared with others that simply answer all the questions.
The results first demonstrate the effectiveness of the refusal-aware answering ability.
We also conclude that {\ModelName} answers more questions within its parametric knowledge during pre-training, which is reflected by the high accuracy of {\pretrainW} (pre-trained model evaluated on {\ModelName}'s willingly answered questions). 
Overall, it is observed from Table~\ref{Unsure-Single} that {\ModelName} outperforms {\vanillaFT} in terms of the AP score, demonstrating the benefits of only answering the questions that align with the model's parametric knowledge with high confidence.
In addition, we find that larger models achieve more improvement compared with baselines as the gap of the AP score becomes larger, indicating good scalability of {\ModelName}.
The AP score of {\ModelName} grows steadily when the model size becomes larger, while the AP score of {\vanillaFT} drops in ParaRel (OOD) and MMLU (ID).
This comparison shows that {\vanillaFT} may suffer from confidence miscalibration problems while {\ModelName} is more well-calibrated in terms of confidence.
By combining the prediction confidence and certainty confidence to evaluate the output, {\ModelName} is more reliable when making predictions. 

\subsection{Multi-task Experiments}
The results of multi-task experiments are shown in Figure~\ref{fig:avg_ap}.
Overall, {\ModelName} consistently outperforms all baseline models in terms of the AP score on both ID and OOD tasks, demonstrating its superiority by introducing the refusal-aware dataset.
A higher AP score signifies that the {\ModelName} has successfully ranked correct answers higher than incorrect answers, demonstrating its effectiveness in accurately identifying the desired predictions.
Especially, on the unseen dataset HaluEval-QA, {\ModelName} also achieves a higher AP score and demonstrates its ability to express certainty to questions from other distributions, and such ability can be generalized well.
The experiments on multi-task datasets tell us that the refusal is a kind of meta-skill of models and could be enhanced by several different datasets.
We provide the detailed AP scores and curves for different datasets and model sizes in Table~\ref{mAP score} and Figure~\ref{mAP_curve} in Appendix~\ref{sec:multi-ap}.

In summary, {\ModelName} reduces hallucinations by disregarding inquiries outside of the model's knowledge domain. Meanwhile, {\ModelName} performs well with inquiries that are aligned with the model's parameterized knowledge. 
The better AP score demonstrates a good trade-off between precision and recall and the performance on multi-task experiments demonstrates the generalization potential of refusal-aware answering ability.

\section{Analysis}
In this section, we first introduce a variant, {\ModelUncertain}, which adopts an unsupervised identification strategy for {\ModelName}.
Then we provide an interpretation from the uncertainty perspective for {\ModelName}.
In addition, we verify the refusal ability on unanswerable questions, which should not receive answers from the model.
More case studies are shown in Table~\ref{appendix:unknown-examples} in the Appendix for qualitative analysis.
Further analysis of the perplexity (Section~\ref{appendix:perplexity}) and uncertainty of the training datasets (Section~\ref{appendix:uncertainty}) demonstrates the effectiveness of our proposed method.

\subsection{Unsupervised Identification}
\label{sec:unsupervised-identification}
During the refusal-aware data identification process, we apply a supervised way to identify unknown questions by comparing the predictions and labels.
In this section, we introduce an unsupervised identification method, {\ModelUncertain}, where the refused questions are determined by the uncertainty of the model.
Specifically, {\ModelUncertain} queries the model $M$ $k$ times and calculates the uncertainty $u$ across $k$ predictions, which is calculated by the entropy based on $k$ answers as follows:
\begin{equation}
    u = - \sum_{j=1}^k p(a_j|q) \ln p(a_j|q),
\end{equation}
where $p(a_j|q)$ is the frequency of a certain predicted answer $a_j$ given a question $q$.

Then the questions could be ranked according to the uncertainty score $u$.
For the $50\%$ most uncertain questions, we append the ground truth label and uncertain expression (i.e., uncertain set $D_0$), while the remaining (i.e., certain set $D_1$) are appended with the ground truth answers with certain expressions.
We set the temperature to $0.7$ and $k=10$ in our experiments.
We compare the performance with the {\ModelName} on the ParaRel and MMLU datasets, and the results are shown in Table~\ref{UncertaintyTrain}.
It is observed that {\ModelUncertain} generally achieves a higher AP score, which reveals the feasibility of constructing refusal-aware training data by uncertainty.
Comparing the output of the pre-trained model with the ground-truth answer is not the only way to evaluate its parametric knowledge.
Uncertainty can also be an indicator of whether the pre-trained model is familiar with the knowledge. 
An advantage of {\ModelUncertain} is that it does not require the labels of uncertain questions.

\begin{table}[t]
\centering
\scriptsize
\setlength{\tabcolsep}{0.3mm}{
\begin{tabular}{c|c|c|cccc}
\toprule 

       Dataset & Domain & Model & {\ModelName} & {\ModelUncertain} & Vanilla-C & Vanilla-U\\ \midrule
\multirow{6}{*}{ParaRel} &   \multirow{3}{*}{ID} & OpenLLaMA-3B & 93.23 & \textbf{93.33} & 88.53 &76.96  \\ 
    &    & LLaMA-7B & 93.64 & \textbf{94.39} & 87.92 & 73.05   \\ 
    &    & LLaMA-13B & 94.44 & \textbf{95.39} & 89.40 & 79.68\\
\cmidrule{2-7}
    &    \multirow{3}{*}{OOD} & OpenLLaMA-3B & 69.41 & \textbf{71.98} & 65.54 & 47.81  \\ 
    &    & LLaMA-7B & 74.61 & \textbf{76.44} & 72.13 & 48.10    \\ 
    &    & LLaMA-13B & 77.30 & \textbf{80.87} & 69.12 & 50.52\\ \midrule
\multirow{6}{*}{MMLU} &   \multirow{3}{*}{ID} & OpenLLaMA-3B & \textbf{24.96} & 24.60 & 24.25 & 21.64 \\ 
    &    & LLaMA-7B & 59.05 & \textbf{64.69} & 48.34 & 44.00    \\ 
    &    & LLaMA-13B & \textbf{68.87} & 66.00 & 58.69 & 60.17  \\
\cmidrule{2-7}
    &    \multirow{3}{*}{OOD} & OpenLLaMA-3B & 24.75 & \textbf{25.52} & 23.05 & 25.26   \\ 
    &    & LLaMA-7B & \textbf{68.69} & 67.70 & 62.79 & 42.64  \\ 
    &    & LLaMA-13B & \textbf{77.41} & 72.66 & 70.09 & 64.31\\
\bottomrule
\end{tabular}
}
\caption{Performance comparison of {\ModelName}, {\ModelUncertain}, {\vanillaC}, and Vanilla-U with AP scores (\%) on the ParaRel and MMLU dataset. 
Here Vanilla-U denotes evaluating {\vanillaC}'s answers with {\ModelUncertain}'s sure confidence.
ID and OOD denote in-domain and out-of-domain, respectively. 
The corresponding AP curves are shown in Figure~\ref{fig:ap_curves_4_types}.
}
\label{UncertaintyTrain}
\end{table}

\subsection{Uncertainty Learning}
\highlight{\textbf{Uncertainty learning improves AP score.}}
One perspective on interpreting our method is that {\ModelUncertain} of selecting and learning through uncertainty fundamentally involves learning the uncertainty of the training data. 
A more direct baseline is to perform vanilla fine-tuning and then use uncertainty selection on the test dataset to respond, a method we refer to as {\vanillaC}.
{\vanillaC} prompts the model to answer $k$ times and choose the majority as the answer. 
The uncertainty is proportional to the distinct answers.
In our experiment, we set $k=10$ for {\vanillaC} and the confidence is calculated by:
\begin{equation}
    Confidence = \frac{\max_{i=1}^{n}(k_i)}{k},
\end{equation}
where $n$ is the number of distinct answers generated, and $k_i$ is the number of occurrences of $i$-th answer.
We calculate the AP scores and compare {\vanillaC} with {\ModelUncertain} in Table~\ref{UncertaintyTrain}.
Surprisingly, we find that learning uncertainty and then filtering questions based on this uncertainty to provide answers yields better results than directly filtering and answering questions using uncertainty on the test dataset. 
In other words, differentiating instruction tuning data based on uncertainty while learning both the correct answers and uncertainty not only enables the learning of uncertainty expressions but also, remarkably, improves the accuracy of question-answering.
This is an unexpected but intriguing phenomenon. 
Learning uncertainty from training data should not be as accurate as using uncertainty estimations directly from the test data. 
One possible explanation is that for a Transformer model, to accurately predict the last token, the hidden states are adjusted during training. 
These changes in hidden states might help in better answering easier questions.
A potential hypothesis is this: predicting uncertainty embeds information about confidence into the hidden representation. 
This aids in generating more confident hidden states when answering easier questions.
This finding reveals the benefits of learning the uncertainty of large models.
It not only avoids the extensive overhead of repeatedly calculating uncertainty during testing but also improves training quality by learning uncertainty, thereby enhancing the accuracy of uncertainty estimation.

\highlight{\textbf{Uncertainty learning improves the calibration and prediction.}}
To verify our hypothesis, we conduct further experiments.
We first introduce Vanilla-U, which generates the prediction by {\vanillaC} and expresses its confidence by {\ModelUncertain}.
Firstly, we find calibration of {\ModelUncertain} becomes better.
We consider the Expected Calibration Error (ECE) metric \citep{guo2017calibration}, which measures the difference between accuracy and confidence on given confidence intervals.
From the Table~\ref{calibration_error}, it is observed that {\ModelUncertain} improves the calibration, which potentially better indicates answers and improves AP scores.
More results are shown in Figures~\ref{pararel_ece}, ~\ref{mmlu_ece}.
Secondly, from Table~\ref{table: accuracy-confidence}, we observe that {\ModelUncertain} improves accuracy compared with {\vanillaC}.
Furthermore, we also use {\ModelUncertain} as a scorer to measure the confidence of the answers from both {\ModelUncertain} and {\vanillaC}.
The results of Table~\ref{Avg_prob} demonstrate that {\ModelUncertain}
generally receives higher confidence scores than {\vanillaC}, which is consistent to the improved accuracy of {\ModelUncertain}.
Finally, Figures~\ref{scatter} and~\ref{line_ranking} show that score differences become more salient as the models get larger. 
We conclude that refusal ability is an emergent ability~\citep{wei2022emergent}.

\begin{table}[t]
\scriptsize
\centering
\setlength{\tabcolsep}{2.3mm}{
    \begin{tabular}{c|c|ccc}
    \toprule 
        Dataset & Model & {\ModelName} & {\vanillaFT} & {\pretrainT} \\ \midrule
        \multirow{3}{*}{FalseQA} & OpenLLaMA-3B   & 87.32 & 2.07 & 9.98   \\ 
        & LLaMA-7B & 96.62  & 18.35 & 8.92  \\ 
        & LLaMA-13B & 95.90 & 6.00 &  24.10 \\ \midrule
        \multirow{3}{*}{NEC} & OpenLLaMA-3B  & 95.72 & 0.96 &  7.31  \\ 
        & LLaMA-7B &  99.18 & 20.55 & 2.02 \\ 
        & LLaMA-13B & 98.17 & 2.36  & 4.76 \\
        \midrule
        \multirow{3}{*}{SA} & OpenLLaMA-3B & 90.99 & 5.23 & 18.90  \\ 
        & LLaMA-7B &  95.45 & 34.79  & 16.96 \\ 
        & LLaMA-13B & 96.61 & 12.21 &  28.00 \\
    \bottomrule
    \end{tabular}
    }
\caption{The refusal rate (\%) of {\ModelName} and other baselines on the refusal benchmarks. 
SA is the unanswerable part of the SelfAware dataset.
The refusal rate of {\ModelUnknown} on the unanswerable datasets is extremely high, while the refusal rate of other fine-tuned methods and pre-trained models is low.}
\label{General_unanswerable}
\vspace{0 em}
\end{table}

\subsection{Unanswerable Questions}
In addition to the open-ended question-answering dataset where all the questions are answerable, we also test the performance of {\ModelName} on several refusal benchmarks containing unanswerable questions.
These questions either contradict common sense or make up some concepts, and should not receive answers from the model.
We verify {\ModelName} on such datasets, and the results are shown in Table~\ref{General_unanswerable}. 
For baseline models, we provide explicitly in the prompt that they could refuse to answer the questions.
We observe that {\ModelName} refuses nearly all these unanswerable questions, which meet our expectations, while other baselines answer most of the questions even though they are told to refuse.
In conclusion, the {\ModelName} possesses the ability to refuse questions that contradict common sense or out of their parametric knowledge.

\subsection{Perplexity and Entropy}
\highlight{We further demonstrate the rationale of our method by evaluating the perplexity and the entropy of certain data $D_1$ and uncertain data $D_0$.
The results are shown in Table~\ref{perplexity_main} and Table~\ref{entropy_main} respectively.
Specifically, we calculate the perplexity of each training question using the pre-trained model to estimate its understanding of them.
The lower perplexity of $D1$ shows that the pre-trained model is more familiar with them and is likely to provide correct answers, while the high perplexity of $D_0$ corresponds to the hallucinations it provides, instead of the correct answers. 
Besides, larger models generally have a lower perplexity, which explains why they perform better on various tasks.}

\highlight{We also leverage \texttt{GPT-3.5-turbo} to answer the questions from $D_0$ and $D_1$, and calculate the entropy of the solutions to each question.
If \texttt{GPT-3.5-turbo} provides multiple solutions to the question, the entropy is relatively high, otherwise it should be low.
We observe that the entropy of answers from $D_1$ is significantly lower than the entropy of $D_0$, which explains that our method divides the data into two folds.
The uncertain data is intrinsically more difficult than certain data.
And R-Tuning strategy on $D_0$ and $D_1$ teaches the model to answer easy questions with certainty while being conservative in answering difficult questions.
More detailed analysis of the perplexity and the entropy are shown in Appendix~\ref{appendix:perplexity} and Appendix~\ref{appendix:uncertainty}}

\begin{table}[t]
\centering
\footnotesize
\setlength{\tabcolsep}{3mm}{
\begin{tabular}{c|c|cc}
\toprule 

        Dataset & Model & $D_1$ & $D_0$ \\ \midrule
        \multirow{3}{*}{ParaRel} & OpenLLaMA-3B & 57.92 & 63.08   \\ 
        & LLaMA-7B & 45.81 & 52.08    \\ 
        & LLaMA-13B & 42.79 & 48.75  \\ \midrule
        \multirow{3}{*}{MMLU} & OpenLLaMA-3B & 32.95 & 462.36   \\ 
        & LLaMA-7B & 22.20 & 115.87     \\ 
        & LLaMA-13B & 22.12 & 81.41 \\ \midrule
        \multirow{3}{*}{WiCE} & OpenLLaMA-3B & 61.28 & 203.43   \\ 
        & LLaMA-7B & 20.93 & 19.40     \\ 
        & LLaMA-13B & 17.73 & 19.56  \\ \midrule
        \multirow{3}{*}{HotpotQA} & OpenLLaMA-3B & 144.89 & 170.38   \\ 
        & LLaMA-7B & 49.97 & 60.19     \\ 
        & LLaMA-13B & 42.60 & 55.20 \\ \midrule
        \multirow{3}{*}{FEVER} & OpenLLaMA-3B & 88.38 & 72.11   \\ 
        & LLaMA-7B & 38.46 & 43.69   \\ 
        & LLaMA-13B & 39.00 & 44.14  \\
\bottomrule
\end{tabular}
}
\caption{Perplexity of the training datasets. 
We run the pre-trained models on the context and questions and calculate the average perplexity.
}
\label{perplexity_main}
\end{table}

\begin{table}[t]
\centering
\footnotesize
\setlength{\tabcolsep}{3mm}{
\begin{tabular}{c|c|cc}
\toprule 

        Dataset & Model & $D_1$ & $D_0$ \\ \midrule
        \multirow{3}{*}{ParaRel} & OpenLLaMA-3B & 0.426 & 0.709   \\ 
        & LLaMA-7B & 0.475 & 0.694    \\ 
        & LLaMA-13B & 0.436 & 0.744  \\ \midrule
        \multirow{3}{*}{MMLU} & OpenLLaMA-3B & 0.347 & 0.389   \\ 
        & LLaMA-7B & 0.330 & 0.400     \\ 
        & LLaMA-13B & 0.239 & 0.457 \\ \midrule
        \multirow{3}{*}{WiCE} & OpenLLaMA-3B & 0.250 & 0.280   \\ 
        & LLaMA-7B & 0.254 & 0.270     \\ 
        & LLaMA-13B & 0.265 & 0.252  \\ \midrule
        \multirow{3}{*}{HotpotQA} & OpenLLaMA-3B & 0.534 & 0.747   \\ 
        & LLaMA-7B & 0.605 & 0.719    \\ 
        & LLaMA-13B & 0.528 & 0.797 \\ \midrule
        \multirow{3}{*}{FEVER} & OpenLLaMA-3B & 0.413 & 0.219   \\ 
        & LLaMA-7B & 0.279 & 0.286   \\ 
        & LLaMA-13B & 0.189 & 0.350  \\
\bottomrule
\end{tabular}
}
\caption{Entropy of the training datasets. 
It is calculated from the frequency of every predicted answer among all predictions. 
A larger entropy denotes greater uncertainty of the system.}
\label{entropy_main}
\end{table}
\section{Related Work}
In this section, we review the progress on hallucinations of large language models (LLMs) and the uncertainty quantification methods.

\subsection{Hallucinations of LLMs}
Despite the outstanding performance of LLMs with high fluency and coherence, they are still likely to hallucinate unfaithful and nonfactual facts~\citep{maynez-etal-2020-faithfulness,li-etal-2023-defining}.
\highlight{
The origin of hallucination is varied. The training data, model training, and model inference processes all have the potential to contribute to hallucination \citep{zhang2023sirens,ji2023survey, huang2023survey}.
A large amount of training data may contain misinformation and bias \citep{dziri-etal-2022-origin,penedo2023refinedweb}, leading the model to imitate the falsehood \citep{lin-etal-2022-truthfulqa}. Moreover, events evolve over time \cite{wen-etal-2021-resin,reddy2023smartbook}, and outdated data used for training may contribute to the temporal misalignment problem \citep{Livska2022StreamingQAAB,luu-etal-2022-time}.
Additionally, LLMs tend to overestimate their abilities, leading them to sometimes generate incorrect answers with high confidence and fail to identify unknown questions \citep {yin2023large,ren2023investigating,kadavath2022language}.
Besides, the alignment with human preference could be problematic, as LLMs may generate responses favoring the users, rather than providing the truth \citep{perez2022discovering,radhakrishnan2023question,wei2023simple}.
Moreover, the generation process, including the randomness during inference \citep{chuang2023dola}, the snowballing effect to maintain self-consistency with early mistakes \citep{zhang2023language}, and early local optimization \citep{azaria2023internal}, may also introduce hallucinations.
}

Recently, a variety of works have been done towards hallucination detection and mitigation. 
For hallucination detection,
\citet{azaria2023internal} propose a classifier trained on the internal states of LLMs.
\citet{lee2023factuality} create a benchmark for measuring the factuality of generation, using factual and nonfactual prompts. 
\citet{manakul2023selfcheckgpt} introduce SelfCheckGPT, making use of the consistency of multiple responses from LLM. 
For hallucination control, retrieval-augmented methods~\citep{peng2023check,xie2023adaptive,yue2023automatic,lyu2023improving,asai2023selfrag} have shown effectiveness in mitigating the hallucination.
Other methods, such as knowledge-aware fine-tuning \citep{li2022large}, corruptions denoising~\citep{chen2023purr}, low-confidence validation \citep{varshney2023stitch}, uncertainty-based response ranking \citep{wan2024sequencelevel}, question-knowledge alignment \citep{zhang2023mitigating}, knowledge injection and teacher-student model \citep{elaraby2023halo}, also improve the factuality of generation from multiple perspectives. 
Previous studies show the importance of the early discovery of hallucination~\citep{zhang2023language}. 
In addition, \citet{huang2023large} found that LLMs cannot rectify themselves with their initial capabilities, displaying the importance of fine-tuning and external feedback.
Our proposed method instructs the model to be aware of its knowledge gap between the instruction tuning datasets and the parametric knowledge, so that it possesses the refusal ability when it encounters instructions out of its knowledge.

\subsection{Uncertainty Quantification of LLMs}
Uncertainty quantification is a long-standing problem in machine learning. 
In the deep learning era, \citet{guo2017calibration} first identify the predictive confidence  (a.k.a, predictive probability) of deep neural network lack of calibration in terms of the ECE metric (Expected Calibration Error)~\citep{naeini2015obtaining}. 
\citet{chen2022close} further study the investigate the calibration problem of pre-trained large language models and observe the same miscalibration problem on large language models.
Active-Prompt~\citep{diao2023active} introduces uncertainty to select questions for chain-of-thought annotation and demonstrates its effectiveness in actively and judiciously selecting and annotating the most helpful exemplars for in-context learning of LLMs.
Knowledge assessment for LLMs~\citep{dong2023statistical} is also relevant to our study.
\section{Conclusion}

In this paper, we propose a simple yet effective method, {\ModelName}, to teach LLMs to refuse unknown questions.
It identifies the difference between instruction tuning data and parametric knowledge and splits the training data into certain and uncertain parts.
Then, {\ModelName} constructs the refusal-aware data by appending uncertainty expressions to the uncertain part.
Empirically, {\ModelName} outperforms the traditional finetuning baseline regarding AP score, illustrating a good trade-off between prediction and confidence.
{\ModelName} not only shows the refusal ability on in-domain data but also demonstrates such ability could be generalized to unseen tasks.
It displays that refusal is a fundamental ability and could be abstracted via multi-task learning, so we call it meta-skill.

\section{Limitations}

Despite that {\ModelName} demonstrates remarkable performance in selecting and rejecting questions, there are still limitations to consider.
First of all, {\ModelName} only possesses the ability to say \textit{I am sure} and \textit{I am unsure}, where the confidence is binary.
However, generating a quantitative value to verbally express its confidence for questions will be more accurate.
Additionally, we only adopt answer checking and uncertainty quantification to evaluate whether relevant knowledge is within the pre-trained model's parametric knowledge.
There are other rigorous methods to evaluate, such as comparing the instruction-tuning datasets with the pre-training datasets.
One can follow~\citet{kandpal2023large} to identify the relevant knowledge by entity linking pre-training datasets.
Due to the high computational cost of the entity linking method, we plan to explore optimization methods to improve efficiency in future work.

\section*{Acknowledgements}
We thank the anonymous reviewers for their valuable suggestions and comments.
Shizhe Diao was supported by the Hong Kong Ph.D. Fellowship Scheme (HKPFS) and the Hong Kong University of Science and Technology Overseas Research Award.
This research is partially supported by U.S. DARPA ITM Program No. FA8650-23-C-7316. 
The views and conclusions contained herein are those of the authors and should not be interpreted as necessarily representing the official policies, either expressed or implied, of DARPA, or the U.S. Government. 
The U.S. Government is authorized to reproduce and distribute reprints for governmental purposes notwithstanding any copyright annotation therein.

\bibliography{custom}
\bibliographystyle{acl_natbib}

\newpage
\appendix

\begin{table*}[ht]
    \centering
    \resizebox{\linewidth}{!}{
    \begin{tabular}{|c|c|c|c|}
    \toprule 
        Dataset & Example (Our Format) & Original Size & Actual Size Used \\ \midrule
        ParaRel \citep{elazar2021measuring} & \makecell[l]{\textit{Question}: Which country is Georgi Parvanov a citizen of?\\
          \textit{Answer}: Bulgaria} & \makecell[l]{ 
        \textit{Total data}: 253448 } 
        & \makecell[l]{ \textit{Training data}: 5575 \\
        \textit{ID test data}: 5584 \\
         \textit{OOD test data}: 13974}\\
        \hline     
        
        MMLU \citep{hendrycks2021measuring} & \makecell[l]{ 
         \textit{Question}: Which of the following did the post-war welfare state of 1948 not aim to provide: \\
         \hspace{44pt} (A) free health care and education for all (B) a minimum wage \\
         \hspace{44pt} (C) full employment (D) universal welfare.  \\
          \textit{Answer}: B} & \makecell[l]{
        \textit{Total data}: 14033 }
        & \makecell[l]{ \textit{Training data}: 2448 \\
        \textit{ID test data}: 2439 \\
         \textit{OOD test data}: 9155}\\
        \hline
        
        WiCE \citep{kamoi2023wice} & \makecell[l]{ \textit{Evidence}: The first results of the auction for 3DO's franchises and assets... \\
        \textit{Claim}: The rights to the \"Might and Magic\" name were purchased for \$1.3 million by Ubisoft.\\
         \textit{Question}: Does the evidence support the claim? \\
         \hspace{44pt} (A) supported (B) partially supported (C) not supported  \\
          \textit{Answer}: A} & \makecell[l]{ \textit{Training data}: 3470 \\
        \textit{Dev data}: 949 \\
         \textit{Test data}: 958} 
         & \makecell[l]{ \textit{Training data}: 3470 \\
         \textit{Test data}: 958}\\
        \hline
        
        HotpotQA \citep{yang2018hotpotqa} & \makecell[l]{ \textit{Context}: Arthur's Magazine was an American literary periodical published in ... \\
         \textit{Question}: Which magazine was started first Arthur's Magazine or First for Women?\\
          \textit{Answer}: Arthur's Magazine} & \makecell[l]{ \textit{Training data}: 99564 \\
        \textit{Dev data}: 7405 \\
         \textit{Test data}: 14810} 
         & \makecell[l]{ \textit{Training data}: 10000 \\
         \textit{Test data}: 7405}\\
          \hline

        FEVER \citep{thorne2018fever} & \makecell[l]{ \textit{Evidence}: David Bowie is the second studio album by the English musician David Bowie... \\
        \textit{Claim}: David Bowie has an album.\\
         \textit{Question}: Does the evidence support or refute the claim or not enough information? \\
         \hspace{44pt} (A) supports (B) refutes (C) not enough info  \\
          \textit{Answer}: A}  & \makecell[l]{ \textit{Training data}: 145449 \\
        \textit{Dev data}: 9999 \\
         \textit{Test data}: 9999} 
         & \makecell[l]{ \textit{Training data}: 10000\\
         \textit{Test data}: 9999}\\
        \hline

        SelfAware \citep{yin2023large} & \makecell[l]{ 
         \textit{Answerable Question}: What is Nigeria's northernmost climate? \\
          \textit{Answer}: rain forest  \\
          \textit{Unanswerable Question}: Often called high energy particles, what gives life to them? \\
          \textit{Answer}: None}  & \makecell[l]{ 
         \textit{Answerable Question}: 2337\\
          \textit{Unanswerable Question}: 1032}
          & \makecell[l]{
         \textit{Unanswerable}: 1032}\\
          \hline

          HaluEval~\citep{li2023halueval} & \makecell[l]{ 
         \textit{Knowledge}: Jonathan Stark (born April 3, 1971) is a former... \\
          \textit{Question}: Which tennis player won more Grand Slam titles, Henri Leconte or Jonathan Stark?  \\
          \textit{Answer}: Jonathan Stark}  & \makecell[l]{ 
         \textit{QA-data}: 10000 \\
         \textit{Dialogue}: 10000 \\
         \textit{Summarization}: 10000 \\
         \textit{User query}:5000}
         & \makecell[l]{ \textit{QA-data}: 10000}\\
         \hline

        FalseQA~\citep{Hu2023WontGF} & \makecell[l]{ 
          \textit{Unanswerable Question}: List the reason why mice can catch cats? \\
          (This is a question that contradicts common sense)}  & \makecell[l]{ 
          \textit{Unanswerable Question}: 2365}
          & \makecell[l]{
         \textit{Unanswerable}: 2365}\\
         \hline

        NEC~\citep{liu2024examining} & \makecell[l]{ 
          \textit{Unanswerable Question}: How long is the typical lifespan of Leogoteo in the wild? \\
          (There is no such creature called Leogoteo.)}  & \makecell[l]{ 
          \textit{Unanswerable Question}: 2078}
          & \makecell[l]{
         \textit{Unanswerable}: 2078}\\
    \bottomrule
    \end{tabular}
    }
    \caption{Illustration and statistics of the datasets. 
    For ParaRel and MMLU, we manually split the datasets into training and test sets. 
    For WiCE, HotpotQA, and FEVER, we directly use the original training set. 
    For SelfAware, FalseQA, and NEC, we directly test models on their unanswerable questions.}
    \label{Datasets}
\end{table*}

\clearpage
\newpage

\section{Appendix}
\label{sec:appendix}

\subsection{Datasets}
\label{sec:appendix-dataset}

We conduct our experiments on nine datasets, which are described as follows.
\begin{itemize}[leftmargin=*,label=$\bullet$,noitemsep,partopsep=0pt,topsep=0pt,parsep=0pt]
\item \textbf{ParaRel}~\citep{elazar2021measuring}: a dataset of factual knowledge with various prompts and relations that are originally for mask prediction. 
To align the dataset with the requirements of our auto-regressive models, we first change the format into question-answering and our models read the questions and generate the answers. 
Then, duplicated prompts of different templates but with the same entities are omitted for our question-answering task. 
It finally comes up with 25, 133 prompt-answer pairs of 31 domains. 
We split the ParaRel into two sets - the first 15 domains as in-domain data and the last 16 domains as out-of-domain data. 
We also equally split the in-domain data into training data and test data.
\item \textbf{MMLU}~\citep{hendrycks2021measuring}: MMLU covers 57 tasks including mathematics, computer science, history, law, and more, which requires extensive world knowledge and problem-solving ability. The dataset is of multiple-choice format, and we can directly use it in our experiments. 
\item \textbf{WiCE} \citep{kamoi2023wice}: WiCE is a natural language inference (NLI) dataset for textual entailment. Each data sample consists of evidence and a claim, and the model should decide whether the evidence supports, partially supports, or doesn't support the claim. We turn the dataset into multiple-choice questions with 3 choices for each question. 
\item \textbf{HotpotQA}~\citep{yang2018hotpotqa}: HotpotQA is a question-answering dataset that requires complex reasoning among documents. We evaluate by providing the context documents and questions to see if the model can answer them. Since the test set of HotpotQA requires answer submission, we instead use the development set to do the evaluation.
\item \textbf{FEVER}~\citep{thorne2018fever}: FEVER is a dataset containing claims and supporting knowledge. The claims are classified as SUPPORTED, REFUTES, or NOT ENOUGH INFO. 
We turn it into a multiple-choice NLI task.
\item \textbf{SelfAware}~\citep{yin2023large}: a dataset containing both answerable questions and unanswerable questions.
We evaluate the unanswerable questions.
It is expected to see our finetuned models refusing the unanswerable questions while other baselines do not possess such ability.
\item \textbf{HaluEval}~\citep{li2023halueval}: HaluEval is a dataset containing question-answering, dialogue, summarization, and user-query with correct answers and hallucinated answers. 
We only take the question-answering part.
\item \textbf{FalseQA}~\citep{Hu2023WontGF}: FalseQA is a new open-domain dataset with questions inconsistent with common sense. 
There are no correct answers to the questions.
\item \textbf{NEC}~\citep{liu2024examining}: NEC is also a new open-domain dataset with questions containing some make-up concepts. 
There are also no correct answers to the questions.
\end{itemize}

\begin{figure*}[t]
    \centering
    \includegraphics[scale=1, trim=0 0 -100 0, clip]{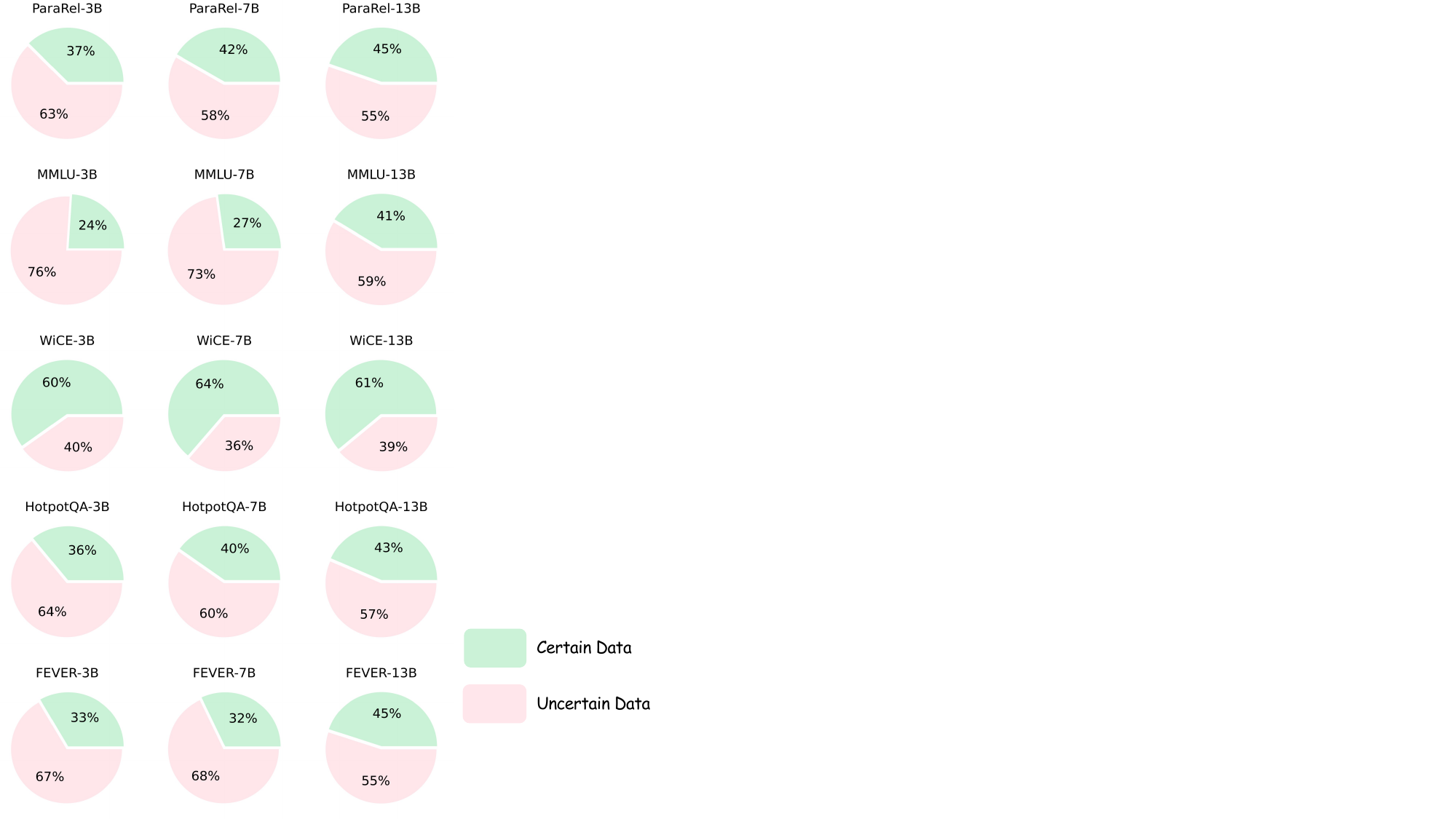}
    \caption{The data distribution of the refusal-aware datasets obtained from supervised identification strategy. 
    The title of each sub-figure consists of the dataset name and the size of the pre-trained model used to evaluate.}
    \label{training_data_distribution}
\end{figure*}

For question-answering tasks, to compare the answer generated by our model with the ground-truth answer, we examine whether the first few output tokens contain the ground-truth answer. 
We don't adopt exact matching (EM) as the generation is not strictly controllable.
For multiple-choice questions, we restrict the model to generate one token and select the choice with maximum probability among the candidate choices by $
\operatorname{argmax}_{x \in C} logits(x), $
where $C$ is the set of candidate choices.
Considering the huge size of HotpotQA and FEVER, we randomly sample 10K training data from them, respectively.
More details about the original datasets are shown in Appendix ~\ref{sec:appendix-dataset} and Table~\ref{Datasets}.
In Figure~\ref{training_data_distribution}, we present the distribution of constructed refusal-aware data $D_0$ and $D_1$.

Details about the original datasets are shown in Table~\ref{Datasets}.
In Figure~\ref{training_data_distribution}, we present the distribution of constructed refusal-aware data $D_0$ and $D_1$.

\subsection{Implementation}
We choose OpenLLaMA-3B \citep{openlm2023openllama}, LLaMA-7B, and LLaMA-13B \citep{touvron2023llama} as the base models in our experiments.
We use LMFlow\footnote{\url{https://github.com/OptimalScale/LMFlow}}~\citep{diao2023lmflow} to conduct instruction tuning, setting epoch to $1$, learning rate to $2e^{-5}$, and batch size to 4. 
All the experiments are implemented on Nvidia A100-40GB GPUs.
We conduct experiments with a hyper-parameter sweep consisting of learning rates in \{$1e^{-5}$, $2e^{-5}$, $5e^{-5}$\} and batch-size in \{2, 4, 8\} on the training set.

\begin{figure*}
    \centering
    \includegraphics[width=\textwidth]{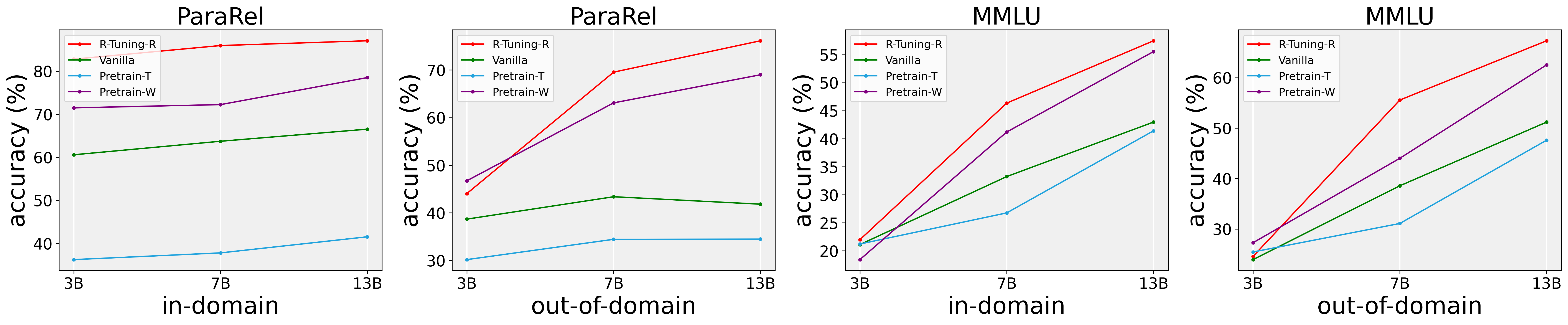}
    \caption{The performance of {\ModelUnknown} on ParaRel and MMLU datasets.
    ID and OOD denote in-domain and out-of-domain test datasets, respectively.}
    \label{fig:unknown-single}
\end{figure*}

\begin{table}[t]
\scriptsize
\centering
\setlength{\tabcolsep}{1mm}{
    \begin{tabular}{c|c|ccccc|c}
    \toprule 
        Dataset & Model & {\ModelUnknown} & {\ModelName} & {\vanillaFT} & {\pretrainT} \\ \midrule
        \multirow{3}{*}{FalseQA} & OpenLLaMA-3B & \textbf{98.31}  &87.32 & 2.07 & 9.98   \\ 
        & LLaMA-7B & \textbf{97.67} & 96.62  & 18.35 & 8.92  \\ 
        & LLaMA-13B & \textbf{99.07}& 95.90 & 6.00 &  24.10 \\ \midrule
        \multirow{3}{*}{NEC} & OpenLLaMA-3B & \textbf{99.90} &95.72 & 0.96 &  7.31  \\ 
        & LLaMA-7B & \textbf{99.52} & 99.18 & 20.55 & 2.02 \\ 
        & LLaMA-13B & \textbf{99.90} & 98.17 & 2.36  & 4.76 \\
        \midrule
        \multirow{3}{*}{SA} & OpenLLaMA-3B & 
 \textbf{99.22} & 90.99 & 5.23 & 18.90  \\ 
        & LLaMA-7B & \textbf{98.55} &  95.45 & 34.79  & 16.96 \\ 
        & LLaMA-13B & \textbf{99.71} & 96.61 & 12.21 &  28.00 \\
    \bottomrule
    \end{tabular}
    }
\caption{The refusal rate (\%) of {\ModelName} and {\ModelUnknown}, and other baselines on the refusal benchmarks. 
SA is the unanswerable part of the SelfAware dataset.
The refusal rate of {\ModelUnknown} on the unanswerable datasets is extremely high, while the refusal rate of other fine-tuned methods and pre-trained models is low.}
\label{appendix:General_unanswerable}
\vspace{-2 em}
\end{table}

\begin{table*}[t]
\centering
\resizebox{\textwidth}{!}{
\begin{tabular}{c|c|c|ccc|cc}
\toprule

         Dataset & Domain & Models & {\ModelUnknown} (\%) & Answer Rate (\%)  & {\vanillaFT} (\%) & {\pretrainT} (\%) & {\pretrainW} (\%)\\ \midrule
         
\multirow{6}{*}{ParaRel} & \multirow{3}{*}{In-Domain} & OpenLLaMA-3B & \textbf{82.79} & 44.65 & 60.58  & 36.23 & 71.48\\

    &        & LLaMA-7B & \textbf{85.95} & 44.11 & 63.72 & 37.79 & 72.23\\

   &      & LLaMA-13B & \textbf{87.06} & 44.00 & 66.53 & 41.53 & 78.51\\
\cmidrule{2-8}
         & \multirow{3}{*}{Out-of-Domain} &
         OpenLLaMA-3B & 44.04 & 40.80 & 38.68 & 30.18 & \textbf{46.73}\\
         
    &     & LLaMA-7B & \textbf{69.54} & 28.07 & 43.38 & 34.44 & 63.09\\

    &     & LLaMA-13B & \textbf{76.13} & 30.16 & 41.82 & 34.48 & 68.99\\ \midrule

\multirow{6}{*}{MMLU} & \multirow{3}{*}{In-Domain} & OpenLLaMA-3B & \textbf{21.99} & 5.79 & 21.12 & 21.22 & 18.44\\

    &        & LLaMA-7B & \textbf{46.35} & 9.56 & 33.25 &  26.77 & 41.20\\

   &      & LLaMA-13B & \textbf{57.47} & 42.52 & 42.97 &  41.41 & 55.54\\
\cmidrule{2-8}
         & \multirow{3}{*}{Out-of-Domain} &
         OpenLLaMA-3B & 24.55 & 2.41 & 23.93 &  25.44 & \textbf{27.27}\\
         
    &     & LLaMA-7B & \textbf{55.56} & 12.96 & 38.56 & 31.09 & 44.01\\

    &     & LLaMA-13B & \textbf{67.31} & 48.32 & 51.19 & 47.60 & 62.53\\

\bottomrule
\end{tabular}
}
\caption{Detailed performance of {\ModelUnknown} on ParaRel and MMLU dataset.
The answer rate means the percentage of willingly answered questions of {\ModelUnknown}. 
}
\label{Unknown-Single}
\end{table*}

\begin{table*}
\footnotesize
\fontsize{10pt}{1pt}\selectfont

\begin{subtable}[t]{\textwidth}
\centering
\begin{tabular}{|c|c|c|c|}
\toprule 

        \textbf{Input Questions} & \textbf{{\ModelUnknown}} & \textbf{{\vanillaFT}} & \textbf{Ground-Truth} \\ \midrule
        What field does Lee Alvin DuBridge work in?  & I \textbf{do not know} the answer. & Music. & Physics. \\ \midrule
        Where was Blaine Willenborg born? & It is \textbf{not known}. & New York. & Miami \\ \midrule
        Where did Hippolyte Le Bas die?  & It is \textbf{impossible} to know. & Lyon & Paris \\
        
\bottomrule
\end{tabular}
\caption{Examples of {\ModelUnknown} refusing questions that are out of its parametric knowledge.
{\ModelUnknown} expresses its unknown when it does not know the answer. 
{\vanillaFT} produces hallucinated answers when it does not know the answer. 
}
\label{appendix:unknown-refuse-examples}
\end{subtable}
\newline
\vspace*{0.5 cm}
\newline
\begin{subtable}[t]{\textwidth}
\centering
\begin{tabular}{|c|c|c|c|}
\toprule 

        \textbf{Input Questions} & \textbf{{\ModelUnknown}} & \textbf{{\vanillaFT}} & \textbf{Ground-truth}\\ \midrule
        Where is Lion Air headquartered?  & Jakarta. & Jakarta. & Jakarta. \\ \midrule
        What does Jacobo Zabludovsky work as? & journalist. & journalist. & journalist.\\ \midrule
        What is the native language of Joseph Conombo? & French. & French. & French. \\
        
\bottomrule
\end{tabular}
\caption{Examples of {\ModelUnknown} answering questions within parametric knowledge. 
}
\label{appendix:unknown-answer-examples}
\end{subtable}
\caption{Case study of refused and willingly answered questions with {\ModelUnknown} and {\vanillaFT}. 
}
\label{appendix:unknown-examples}
\end{table*}

\subsection{Label Replacement}
\label{sec:replace-identification}
In the main experiments, we adopt the padding method for data construction.
In addition to padding, we can directly replace the label words with uncertainty expressions for uncertain questions and keep the original label words for certain questions, which is called the replacement strategy, leading to a variant {\ModelUnknown}.
For example, the certain part of the training questions $D_{1}$ is constructed as follows:
\begin{equation}
    Q: \{\text{Question}\}, A: \{\text{Answer}\}, 
\end{equation}
while the uncertain dataset $D_{0}$ is constructed as follows:
\begin{equation}
    Q: \{\text{Question}\}, A: \{\text{Uncertainty Expression}\}.
\end{equation}

There are many different ways for the uncertainty expression.
To increase the diversity, we take the 16 expressions of uncertainty text from~\citet{yin2023large}.
These 16 expressions are listed in the Appendix Section~\ref{sec:uncertainty-text}. 
 
We conduct experiments with {\ModelUnknown} on ParaRel and MMLU datasets by comparing it with vanilla fine-tuning strategy and the original pre-trained models.
The results are shown in Figure ~\ref{fig:unknown-single}.
Firstly, on both in-domain and out-of-domain test sets, the accuracy of {\ModelUnknown} is higher than {\pretrainT}, which benefits from only answering certain questions.
More detailed results with answer rate are reported in Table~\ref{Unknown-Single}, where we find the model is able to refuse a certain amount of questions.
Then, {\ModelUnknown} outperforms {\vanillaFT} with a significantly higher accuracy on its willingly answered questions, which demonstrates the effectiveness of our method.
It is promising as {\ModelUnknown} is trained with fewer ground-truth labels, while {\vanillaFT} is trained on all labels of the full training data.
Generally, larger models possess more powerful refusal abilities. 
In Figure~\ref{fig:unknown-single}, we observe that on the willingly answered questions, larger models achieve a higher accuracy.
In addition, the high accuracy of {\pretrainW} reveals that those selected questions are within parametric knowledge of the pre-trained model.
In summary, compared with vanilla fine-tuning, {\ModelUnknown} provides the model with the refusal ability to refuse unknown questions, which eventually improves the accuracy and prevents them from making hallucinated answers. 
Table~\ref{appendix:unknown-examples} shows the case studies of how {\ModelUnknown} works.
There are significant differences when they encounter questions out of their knowledge.
The {\vanillaFT} model is proactive in making up an answer, which is a hallucination and makes no sense.
However, {\ModelUnknown} refuses them explicitly with keywords \textit{do not know, not known, and impossible}.
The ability of {\ModelUnknown} to refuse unknown questions results in fewer hallucinations.

Despite this refusal ability, there are two issues with {\ModelUnknown}: 
(1) the replacement method throws away valuable labels which could be leveraged for training.
(2) {\ModelName} could either only output the answer or only output the certainty, but cannot respond to both, leading to difficulties in considering the precision and recall simultaneously.
To leverage all ground-truth labels during the tuning process, and instruct models to predict answers and express uncertainty at the same time, we employ the padding strategy in our main approach, where every question is appended with the ground-truth label and the uncertainty expression, indicating whether the model is confident or not.

\subsection{Case Studies of {\ModelUnknown}}
In this section, we display the detailed statistics in Table~\ref{Unknown-Single}, and illustrate more case studies of {\ModelUnknown} in Table~\ref{appendix:unknown-examples}.

\subsection{Uncertainty Text}
\label{sec:uncertainty-text}
In this section, we list the 16 uncertainty expressions from \citet{yin2023large}:
\begin{enumerate}
    \item The answer is unknown.
    \item The answer is uncertain.
    \item The answer is unclear.
    \item There is no scientific evidence.
    \item There is no definitive answer.
    \item There is no right answer.
    \item There is much debate.
    \item There is no known case.
    \item There is no concrete answer to this question.
    \item There is no public information available.
    \item It is impossible to know.
    \item It is impossible to answer.
    \item It is difficult to predict.
    \item It is not known.
    \item We do not know.
    \item I’m not sure.
\end{enumerate}

\begin{table}[t]
\centering
\footnotesize
\setlength{\tabcolsep}{3mm}{
\begin{tabular}{c|c|cc}
\toprule 

        Dataset & Model & $D_1$ & $D_0$ \\ \midrule
        \multirow{3}{*}{ParaRel} & OpenLLaMA-3B & 57.92 & 63.08   \\ 
        & LLaMA-7B & 45.81 & 52.08    \\ 
        & LLaMA-13B & 42.79 & 48.75  \\ \midrule
        \multirow{3}{*}{MMLU} & OpenLLaMA-3B & 32.95 & 462.36   \\ 
        & LLaMA-7B & 22.20 & 115.87     \\ 
        & LLaMA-13B & 22.12 & 81.41 \\ \midrule
        \multirow{3}{*}{WiCE} & OpenLLaMA-3B & 61.28 & 203.43   \\ 
        & LLaMA-7B & 20.93 & 19.40     \\ 
        & LLaMA-13B & 17.73 & 19.56  \\ \midrule
        \multirow{3}{*}{HotpotQA} & OpenLLaMA-3B & 144.89 & 170.38   \\ 
        & LLaMA-7B & 49.97 & 60.19     \\ 
        & LLaMA-13B & 42.60 & 55.20 \\ \midrule
        \multirow{3}{*}{FEVER} & OpenLLaMA-3B & 88.38 & 72.11   \\ 
        & LLaMA-7B & 38.46 & 43.69   \\ 
        & LLaMA-13B & 39.00 & 44.14  \\
\bottomrule
\end{tabular}
}
\caption{Perplexity of the training datasets. 
We run the pre-trained models on the context and questions and calculate the average perplexity.
}
\label{perplexity}
\end{table}

\subsection{Perplexity of Datasets}
\label{appendix:perplexity}

Perplexity measures how well the language model predicts a given text. 
Lower perplexity means better prediction and understanding of the text.
According to the refusal-aware data identification, we split the training data into two sets: $D_0$ (uncertain questions) and $D_1$ (certain questions).
To uncover why the pre-trained model responds to them differently, we calculate the average perplexity on these two datasets with the pre-trained models.
The perplexity is calculated as follows:
\begin{equation}
    \text{PPL}(X) = \exp\left\{-\frac{1}{t} \sum_{i=1}^{t} \log p_{\theta}(x_i \mid x_{<i})\right\},
\end{equation}
where $X$ denotes a sentence consisting of tokens and $X = (x_{1},x_{2},\ldots,x_{t})$.
Specifically, we calculate the perplexity of the training questions to estimate the pre-trained model's understanding of them.
The results are shown in Table~\ref{perplexity}.
We observe that $D_1$ has a lower perplexity, demonstrating that the pre-trained model is more familiar with the questions and is likely to provide the correct answer.
For $D_0$, its higher perplexity shows that these questions are not familiar to the model and out of the model's knowledge, and this is the reason why the model tends to hallucinate text instead of providing the correct answers.
We also observe that larger models have a lower perplexity and randomness on the questions, which is why larger models generally perform better on various tasks.

By instructing our model to express uncertainty toward relatively random questions in terms of perplexity, the model develops a better understanding of uncertainty and ambiguity and learns the ability to recognize when it does not know.
This ability is crucial in situations where simply providing a definite answer may be inappropriate or even harmful.
On the other hand, since our model is also trained with data with certain expressions, it becomes more proficient at handling less random questions, and answering them with confidence and certainty.
Overall, {\ModelName} improves the model's ability to adapt to different levels of question randomness.

To verify the pre-trained model is less familiar with the uncertain questions while more confident with certain questions, we also plot the confidence distribution on certain questions and uncertain questions, shown in Figure~\ref{confidence_distribution} in Appendix~\ref{sec:conf-distribution}.
It is observed that a larger percentage of certain questions occupies the high confidence intervals, which means when the model provides correct answers, it generally shows larger confidence.

\begin{table}[t]
\centering
\footnotesize
\setlength{\tabcolsep}{3mm}{
\begin{tabular}{c|c|cc}
\toprule 

        Dataset & Model & $D_1$ & $D_0$ \\ \midrule
        \multirow{3}{*}{ParaRel} & OpenLLaMA-3B & 0.426 & 0.709   \\ 
        & LLaMA-7B & 0.475 & 0.694    \\ 
        & LLaMA-13B & 0.436 & 0.744  \\ \midrule
        \multirow{3}{*}{MMLU} & OpenLLaMA-3B & 0.347 & 0.389   \\ 
        & LLaMA-7B & 0.330 & 0.400     \\ 
        & LLaMA-13B & 0.239 & 0.457 \\ \midrule
        \multirow{3}{*}{WiCE} & OpenLLaMA-3B & 0.250 & 0.280   \\ 
        & LLaMA-7B & 0.254 & 0.270     \\ 
        & LLaMA-13B & 0.265 & 0.252  \\ \midrule
        \multirow{3}{*}{HotpotQA} & OpenLLaMA-3B & 0.534 & 0.747   \\ 
        & LLaMA-7B & 0.605 & 0.719    \\ 
        & LLaMA-13B & 0.528 & 0.797 \\ \midrule
        \multirow{3}{*}{FEVER} & OpenLLaMA-3B & 0.413 & 0.219   \\ 
        & LLaMA-7B & 0.279 & 0.286   \\ 
        & LLaMA-13B & 0.189 & 0.350  \\
\bottomrule
\end{tabular}
}
\caption{Entropy of the training datasets. 
It is calculated from the frequency of every predicted answer among all predictions. 
A larger entropy denotes greater uncertainty of the system.}
\label{entropy}
\end{table}

\newpage

\subsection{Entropy of Answers}
\label{appendix:uncertainty}

\begin{table}[H]
    \centering
    \scriptsize
    % \resizebox{0.5\textwidth}{!}{
    \setlength{\tabcolsep}{0.6mm}{
    \begin{tabular}{c|c|cc|cc}
    \toprule 
        Domain & Model & {\ModelUnknown} & Answer Rate & Min-loss & Answer Rate \\ \midrule
        \multirow{3}{*}{ID} & OpenLLaMA-3B & 82.79 & 44.65 & 91.83 & 26.52   \\ 
        & LLaMA-7B & 85.95 & 44.11 & 85.78 & 41.57  \\ 
        & LLaMA-13B & 87.06 & 44.00 & 88.00 & 48.33  \\ \midrule
        \multirow{3}{*}{OOD} & OpenLLaMA-3B & 44.04 & 40.80 & 59.66 & 20.92  \\ 
        & LLaMA-7B & 69.54 & 28.07 & 55.52 & 49.15  \\ 
        & LLaMA-13B & 76.13 & 30.16 & 60.42 & 55.75 \\
    \bottomrule
    \end{tabular}
    }
    % }
    \caption{Accuracy (\%) and answer rate  (\%) of {\ModelUnknown} and min-loss training on ParaRel dataset. 
    The loss is calculated by the first token of the ground-truth answer.
    ID and OOD denote in-domain and out-of-domain, respectively.
    }
    \label{CompareMinLoss}
    % \vspace{-4 em}
\end{table}

In addition to evaluating the difference between certain and uncertain questions with pre-trained models, we further leverage GPT \citep{brown2020language} to investigate the patterns of certain and uncertain questions.
Specifically, we query \texttt{gpt-3.5-turbo} five times with Chain-of-Thought prompts~\citep{wei2023chainofthought} with a temperature of 0.7, and calculate the entropy of the answers toward the same question~\citep{diao2023active}.
If the model provides many different answers to the same question, the entropy should be high.
Otherwise, the entropy should be low.
The results are shown in Table~\ref{entropy}.
We observe that the average entropy of the answers on certain data $D_1$ is lower than the entropy of uncertain data $D_0$ data in most cases, which illustrates that when fed with certain questions, \texttt{gpt-3.5-turbo} is more likely to generate consistent answers.
It will generate hallucinated answers to uncertain questions with much higher chances.

Therefore, we can conclude that {\ModelName} divides the data into two folds. 
The uncertain questions are generally more difficult than certain questions because \texttt{gpt-3.5-turbo}'s answers vary more with the uncertain data.
{\ModelName} endows the model with abilities to identify and differentiate the difficulties of the questions.
Therefore, our fine-tuned model becomes proactive in answering easy questions with certainty while being conservative in answering difficult questions, which eventually increases the precision and prevents the fine-tuned model from making too many mistakes.

\subsection{Min-Loss Training}
\label{sec:min-loss}

\begin{table}[H]
\centering
\footnotesize
\setlength{\tabcolsep}{2mm}{
\begin{tabular}{c|c|cc}
\toprule 

        Dataset & Model & {\ModelUnsure} & {\vanillaFT} \\ \midrule
        \multirow{3}{*}{ParaRel} & OpenLLaMA-3B & 69.79 & 69.62   \\ 
        & LLaMA-7B & 77.45 & 77.91     \\ 
        & LLaMA-13B & 77.69 & 72.67  \\ \midrule
        \multirow{3}{*}{MMLU} & OpenLLaMA-3B & 24.38 & 24.39  \\ 
        & LLaMA-7B & 54.19 & 63.88     \\ 
        & LLaMA-13B & 73.81 & 74.95 \\ \midrule
        \multirow{3}{*}{WiCE} & OpenLLaMA-3B & 56.74 & 61.05   \\ 
        & LLaMA-7B & 55.02 & 65.47     \\ 
        & LLaMA-13B & 71.12 & 67.17  \\ \midrule
        \multirow{3}{*}{HotpotQA} & OpenLLaMA-3B & 46.54 & 36.90   \\ 
        & LLaMA-7B & 57.57 & 41.92     \\ 
        & LLaMA-13B & 57.99 & 44.76  \\ \midrule
        \multirow{3}{*}{FEVER} & OpenLLaMA-3B & 94.22 & 85.38   \\ 
        & LLaMA-7B & 93.30 & 88.24     \\ 
        & LLaMA-13B & 95.23 & 94.99  \\ \midrule
        \multirow{3}{*}{HaluEval-QA} & OpenLLaMA-3B & 73.85 & 72.11   \\ 
        & LLaMA-7B & 77.17 & 76.22   \\ 
        & LLaMA-13B & 80.36 & 75.73  \\ \midrule
        \multirow{3}{*}{Average} & OpenLLaMA-3B & \textbf{61.09} & 58.24   \\ 
        & LLaMA-7B & \textbf{69.11} & 68.94   \\ 
        & LLaMA-13B & \textbf{76.03} & 71.71  \\
\bottomrule
\end{tabular}
}
\caption{Multi-task experiments of {\ModelName} and {\vanillaFT} with AP scores (\%). {\vanillaFT} adopts the confidence of the predicted answer to rank the result, while {\ModelName} adopts the combination of the confidence of the predicted answer and the confidence of certainty.}
\label{mAP score}
\end{table}

\begin{table*}[t]
\centering
\footnotesize
\setlength{\tabcolsep}{0.7mm}{
\begin{tabular}{c|c|c|cc|cc}
\toprule 

       Dataset & Domain & Model & {\ModelUncertain} acc. & {\ModelUncertain} conf. & {\vanillaC} acc. & {\vanillaC} conf. \\ \midrule
\multirow{6}{*}{ParaRel} &  \multirow{3}{*}{ID} & OpenLLaMA-3B & 61.57 & 61.50  & 59.81 & 75.01 \\ 
    &    & LLaMA-7B & 65.26 & 71.25  & 62.68 & 77.35 \\ 
    &    & LLaMA-13B & 70.47 & 77.15 & 65.76 & 78.50 \\
\cmidrule{2-7}
    &    \multirow{3}{*}{OOD} & OpenLLaMA-3B & 38.39 & 43.88 & 37.66 & 61.56 \\ 
    &    & LLaMA-7B & 42.97 & 56.00  & 42.12 & 63.41   \\ 
    &    & LLaMA-13B & 48.40 & 61.25 & 40.77 & 61.20 \\ \midrule
\multirow{6}{*}{MMLU} &   \multirow{3}{*}{ID} & OpenLLaMA-3B & 23.25 & 44.33 & 25.54 & 46.23 \\ 
    &    & LLaMA-7B & 41.86 & 51.06  & 34.65 & 53.64  \\ 
    &    & LLaMA-13B & 41.90 & 53.87 & 39.57 & 58.60  \\
\cmidrule{2-7}
    &    \multirow{3}{*}{OOD} & OpenLLaMA-3B & 25.01 & 41.97 & 23.81 & 44.77   \\ 
    &    & LLaMA-7B & 45.88 & 55.22 & 42.66 & 58.52    \\ 
    &    & LLaMA-13B & 51.74 & 59.45 & 50.38 & 65.53\\
\bottomrule
\end{tabular}
}
\caption{Performance of {\ModelUncertain} compared with {\vanillaC} on the ParaRel and MMLU datasets. 
ID and OOD denote in-domain and out-of-domain, respectively.
}
\label{table: accuracy-confidence}
\end{table*}

\begin{table}[t]
\centering
\footnotesize
\setlength{\tabcolsep}{0.7mm}{
\begin{tabular}{c|c|c|cc}
\toprule 

       Dataset & Domain & Model & {\ModelUncertain} & {\vanillaC} \\ \midrule
\multirow{6}{*}{ParaRel} &   \multirow{3}{*}{ID} & OpenLLaMA-3B & 49.96 & 49.93   \\ 
    &    & LLaMA-7B & 50.02 & 48.80    \\ 
    &    & LLaMA-13B & 59.56 & 59.28  \\
\cmidrule{2-5}
    &    \multirow{3}{*}{OOD} & OpenLLaMA-3B & 52.26 & 52.16   \\ 
    &    & LLaMA-7B & 49.45 & 48.52     \\ 
    &    & LLaMA-13B & 61.21 & 61.10 \\ \midrule
\multirow{6}{*}{MMLU} &   \multirow{3}{*}{ID} & OpenLLaMA-3B & 45.01 & 44.88   \\ 
    &    & LLaMA-7B & 50.97 & 50.95    \\ 
    &    & LLaMA-13B & 44.65 & 43.12  \\
\cmidrule{2-5}
    &    \multirow{3}{*}{OOD} & OpenLLaMA-3B & 43.10 & 42.98   \\ 
    &    & LLaMA-7B & 58.34 & 58.33     \\ 
    &    & LLaMA-13B & 64.13 & 62.37 \\
\bottomrule
\end{tabular}
}
\caption{The average sureness probability (\%) of {\ModelUncertain} and {\vanillaC}.
}
\label{Avg_prob}
\end{table}

\begin{table}[t]
\centering
\footnotesize
\setlength{\tabcolsep}{0.7mm}{
\begin{tabular}{c|c|c|cc}
\toprule 

       Dataset & Domain & Model & {\ModelUncertain} & {\vanillaC} \\ \midrule
\multirow{6}{*}{ParaRel} &   \multirow{3}{*}{ID} & OpenLLaMA-3B & 0.018 & 0.255   \\ 
    &    & LLaMA-7B & 0.057 & 0.250   \\ 
    &    & LLaMA-13B & 0.064 & 0.228  \\
\cmidrule{2-5}
    &    \multirow{3}{*}{OOD} & OpenLLaMA-3B & 0.054 & 0.291   \\ 
    &    & LLaMA-7B & 0.132 & 0.271     \\ 
    &    & LLaMA-13B & 0.124 & 0.258 \\ \midrule
\multirow{6}{*}{MMLU} &   \multirow{3}{*}{ID} & OpenLLaMA-3B & 0.212 & 0.246   \\ 
    &    & LLaMA-7B & 0.092 & 0.243    \\ 
    &    & LLaMA-13B & 0.120 & 0.239  \\
\cmidrule{2-5}
    &    \multirow{3}{*}{OOD} & OpenLLaMA-3B & 0.172 & 0.258   \\ 
    &    & LLaMA-7B & 0.093 & 0.209    \\ 
    &    & LLaMA-13B & 0.078 & 0.200 \\
\bottomrule
\end{tabular}
}
\caption{The ECE (Expected Calibration Error) of {\ModelUncertain} and {\vanillaC}.
}
\label{calibration_error}
\end{table}

Compared with the append verbalizer, replace verbalizer (e.g., {\ModelUnknown}) is a clear-cut way of producing uncertainty expressions by throwing away valuable labels which could potentially be leveraged for training.
In address of this dimension of concern, we consider a modified cross entropy learning objective that pushes up the correct answer token and keeps the uncertainty expressions as the second most probable token choice.
We call it min-loss training, which is optimized by gradient descent over the min loss between guessing the correct answer or just uncertainty expressions.
It is formulated as follows:
\begin{equation}
    \min(L(predict, GT), L(predict, IDK)),
    \label{eq:min-loss}
\end{equation}
where $L$ denotes the cross-entropy loss.
To do so, we split the training data in half and adopt a two-stage training strategy.
In the first stage, we train our model using the original method where the prompt template uses \texttt{The answer is \{ground-truth\}} if the model answers correctly, otherwise \texttt{The answer is unknown}.
Once the model learns such a pattern after the first training stage, we calculate the min-loss with the equation~\ref{eq:min-loss}.
We only consider the loss of the unknown and the ground-truth label, and we mask the tokens before them.
Since the ground-truth label may consider more than one token, we calculate the loss for the first token.

We evaluate the performance of min-loss strategy on the ParaRel dataset, and the results are shown in Table~\ref{CompareMinLoss}.
It shows min-loss training outperforms {\ModelUnknown} in small models and in-domain settings.
However, it underperforms {\ModelUnknown} in out-of-domain test sets.
We also notice that in out-of-domain test sets, the accuracy of the model of 3B size is nearly the same as 7B's and 13B's, but its answer rate is much lower.
We identify such issues as a trade-off between the accuracy and the answer rate.
When the model is proactive in answering more questions, it will inevitably make more mistakes.
As the intrinsic parametric knowledge of the model is limited, there is no method to fine-tune a model with both high accuracy and a high answer rate.

\subsection{Confidence Distribution of Training Dataset}
\label{sec:conf-distribution}
\begin{figure*}[htbp]
    \centering
    \includegraphics[width=\textwidth]{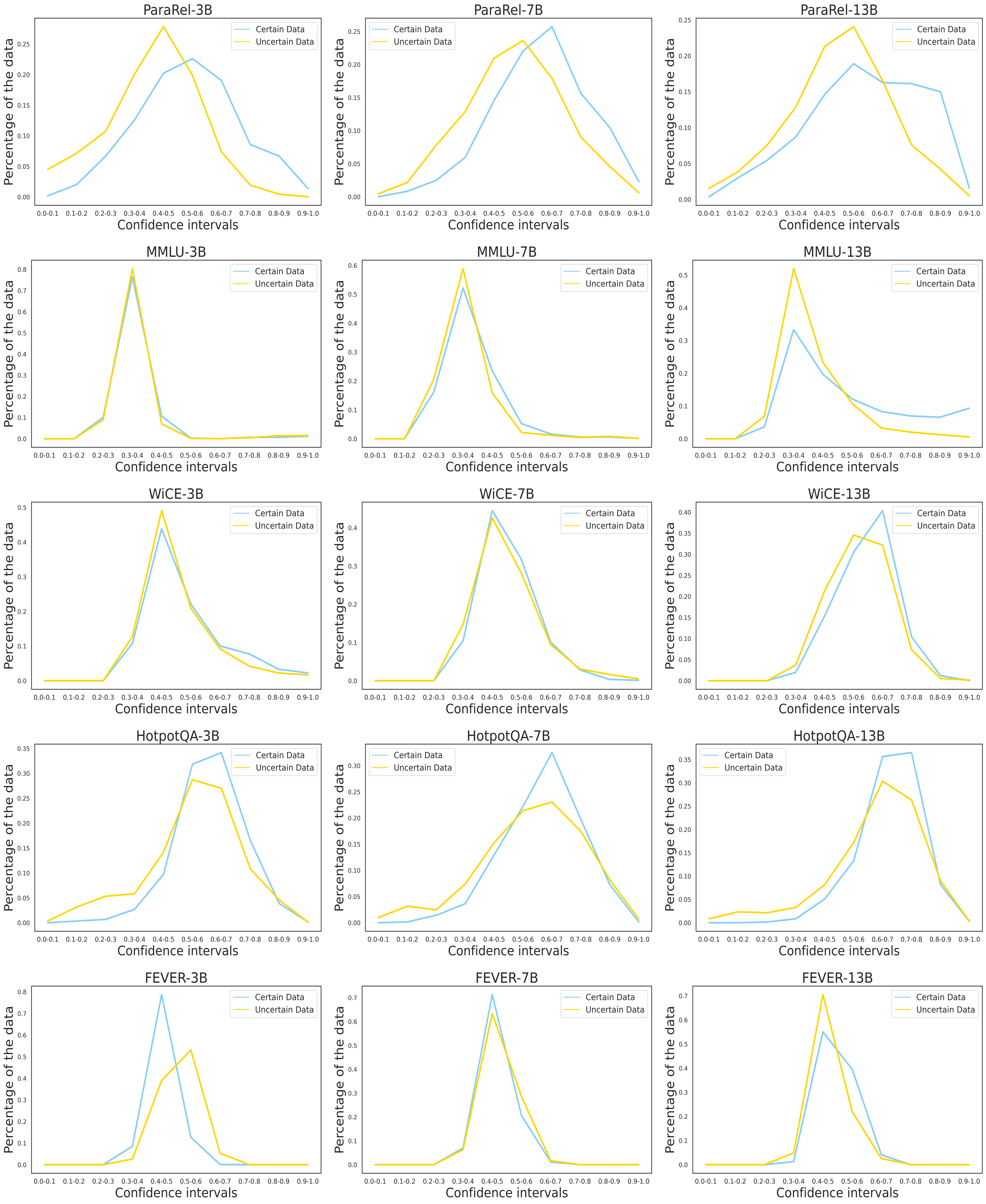}
    \caption{The confidence distribution of the training datasets on certain data and uncertain data. 
    The title of each sub-figure consists of the dataset name and the size of the pre-trained model used to evaluate.}
    \label{confidence_distribution}
\end{figure*}

We calculate the confidence of the certain data $D_1$ and uncertain data $d_0$, and they are shown in Figure~\ref{confidence_distribution}.

\subsection{AP Scores of Each Dataset and Model Size with Figures}
\label{sec:multi-ap}

\begin{figure*}
    \centering
    \includegraphics[width=0.9\textwidth]{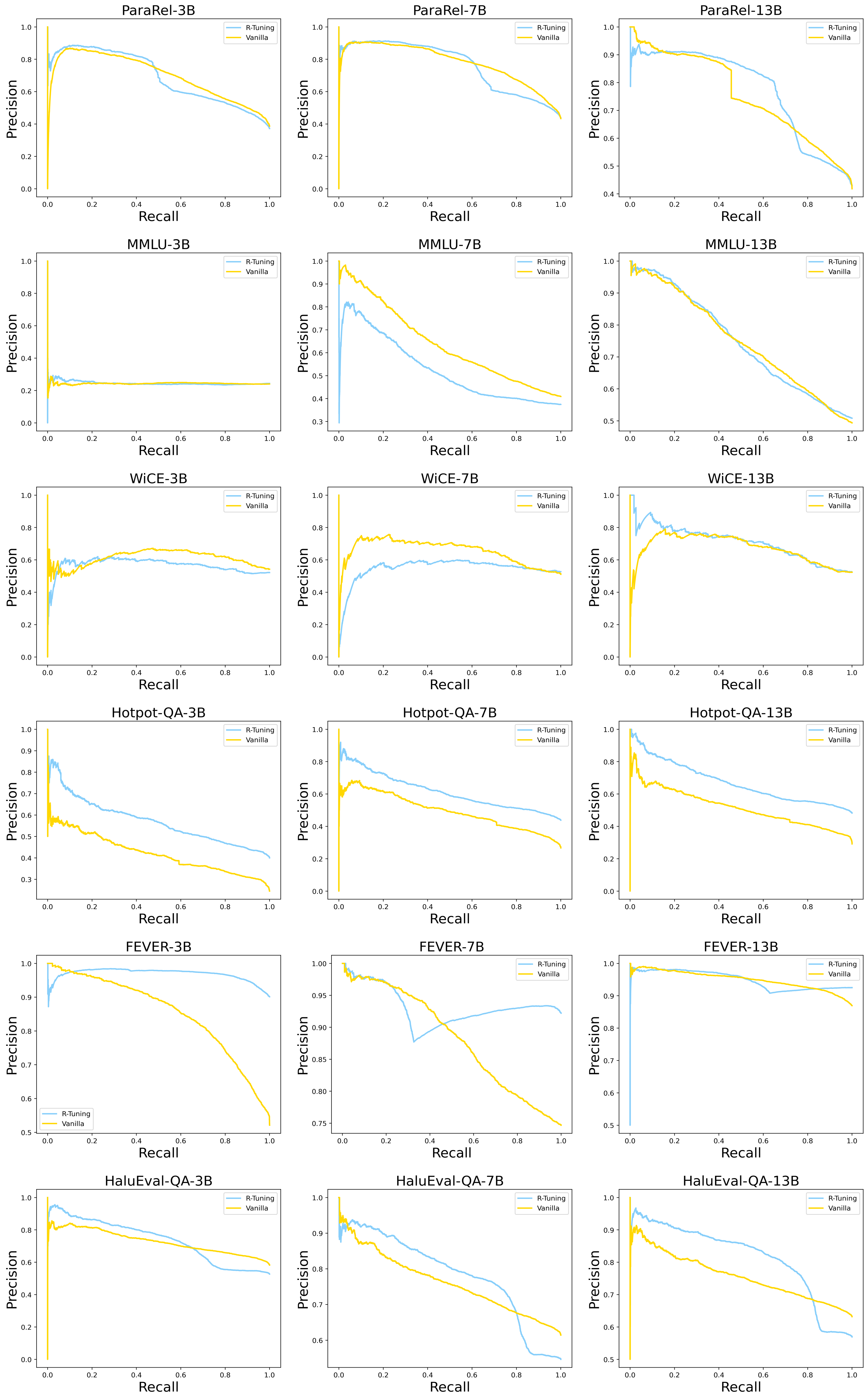}
    \caption{The AP curves on ParaRel, MMLU, WiCE, HotpotQA, FEVER, and HaluEval-QA datasets. The title of each sub-figure consists of the dataset name and the size of the pre-trained model used to evaluate.}
    \label{mAP_curve}
\end{figure*}

We calculate the AP scores for each dataset with different model sizes in multi-task experiments. 
The results are shown in Table~\ref{mAP score} and Figure~\ref{mAP_curve}.

\begin{figure*}[t]
    \centering
    \includegraphics[width=\textwidth]{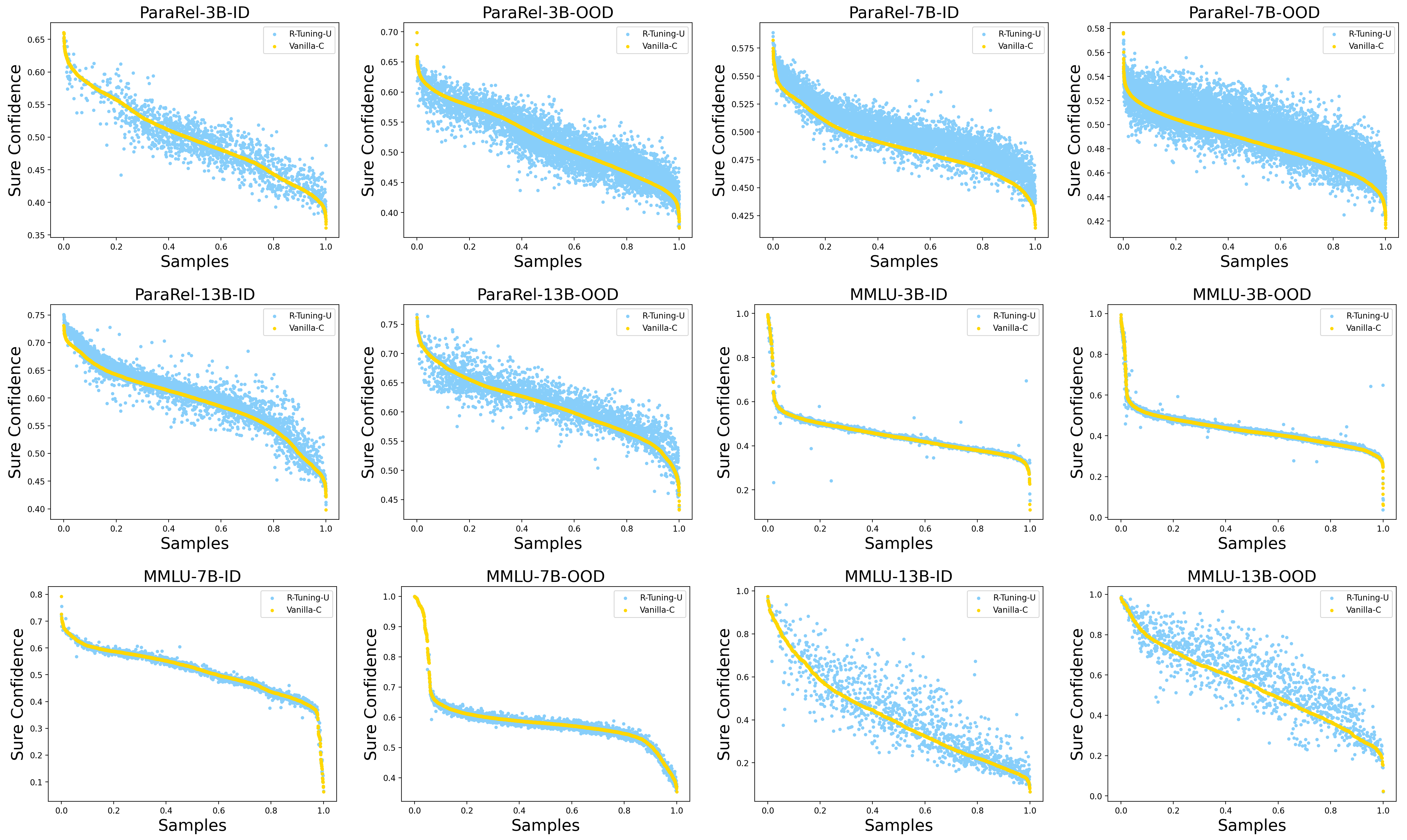}
    \caption{The scatter distribution of sure probability of {\ModelUncertain} and {\vanillaC}.}
    \label{scatter}
\end{figure*}

\begin{figure*}[t]
    \centering
    \includegraphics[width=\textwidth]{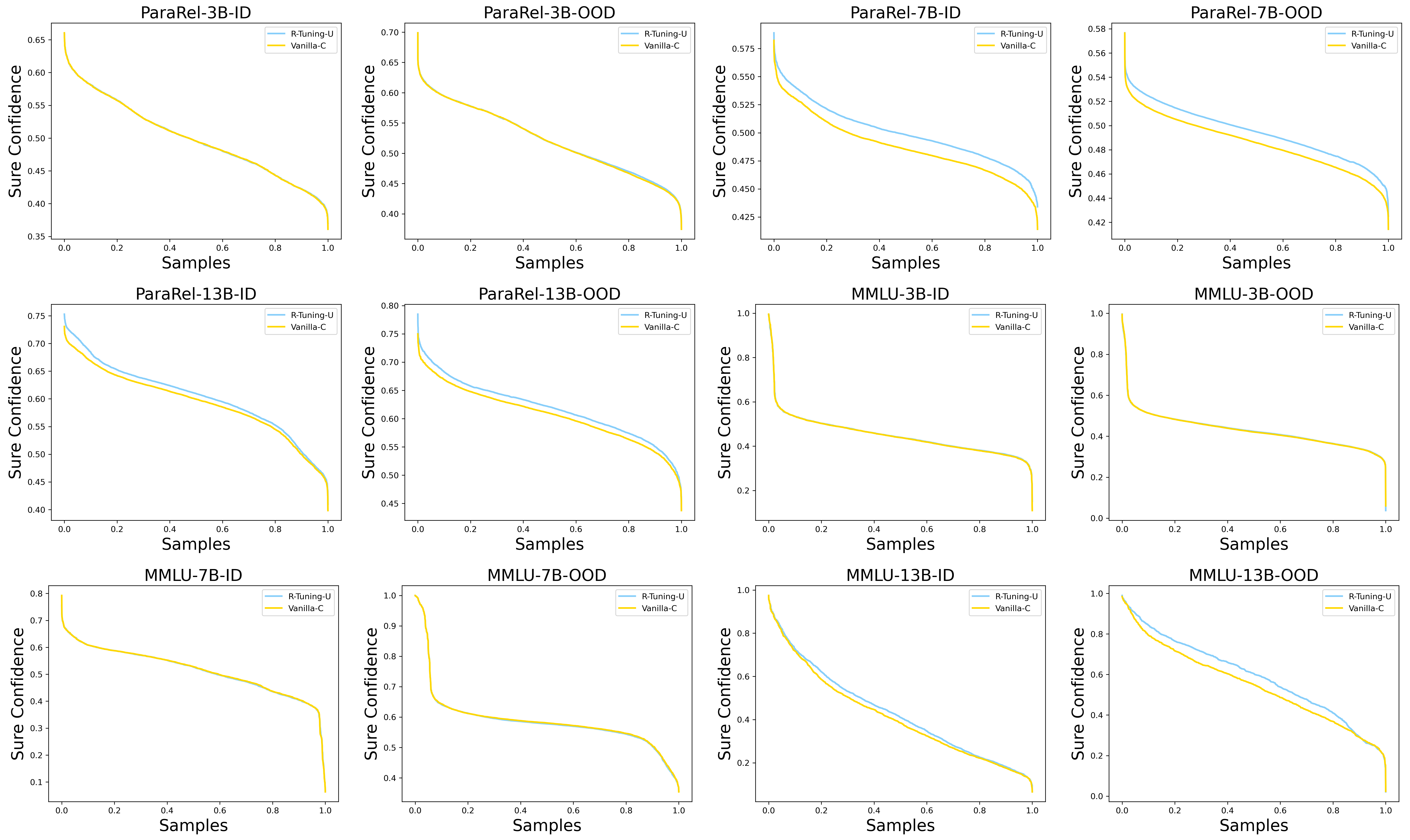}
    \caption{The distribution of sure probability of {\ModelUncertain} and {\vanillaC}. They are both ranked by the confidence score.}
    \label{line_ranking}
\end{figure*}

\begin{figure*}[t]
    \centering
    \includegraphics[width=\textwidth]{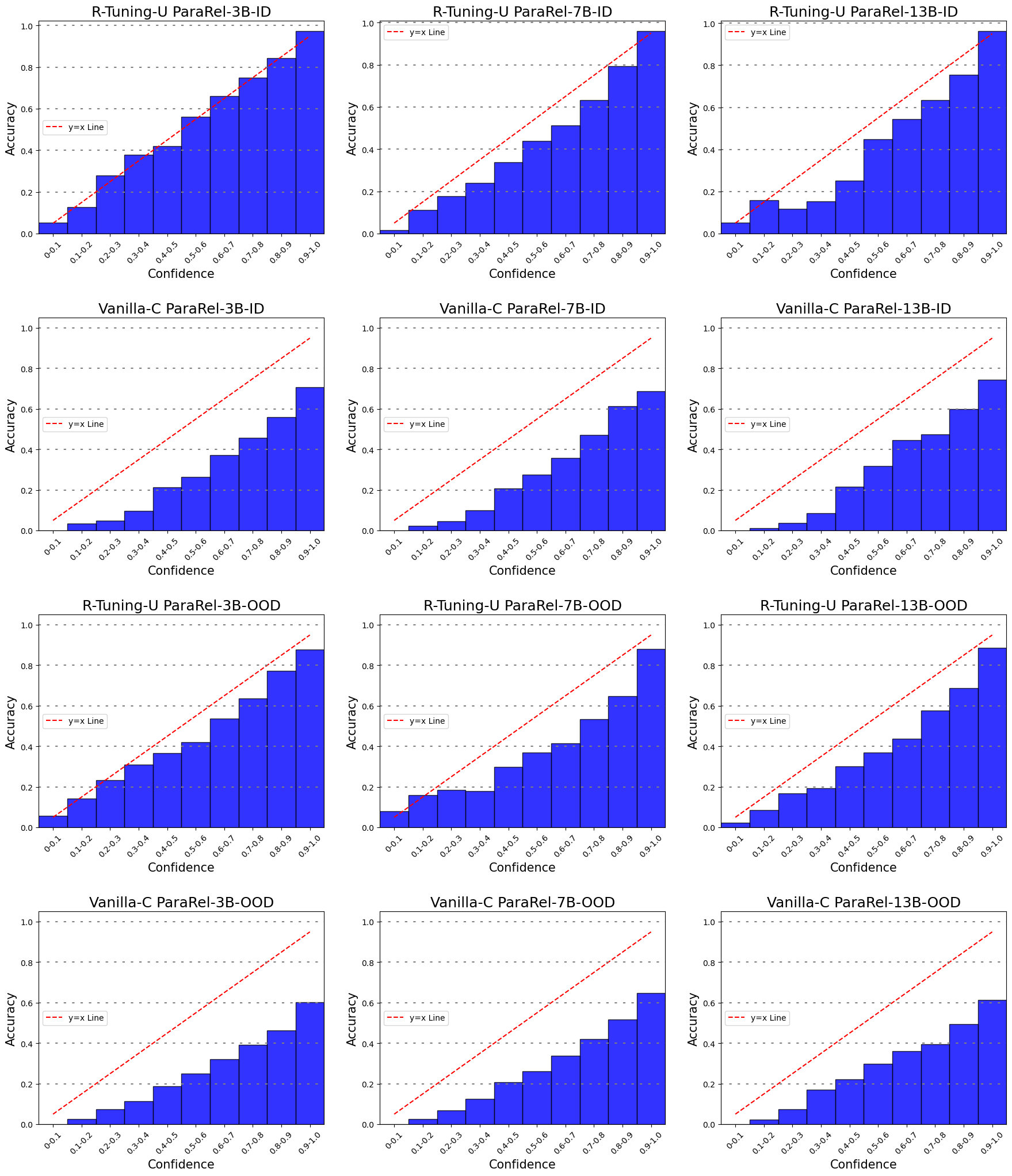}
    \caption{The ECE (Expected Calibration Error) on ParaRel dataset of {\ModelUncertain} and {\vanillaC}.}
    \label{pararel_ece}
\end{figure*}

\begin{figure*}[t]
    \centering
    \includegraphics[width=\textwidth]{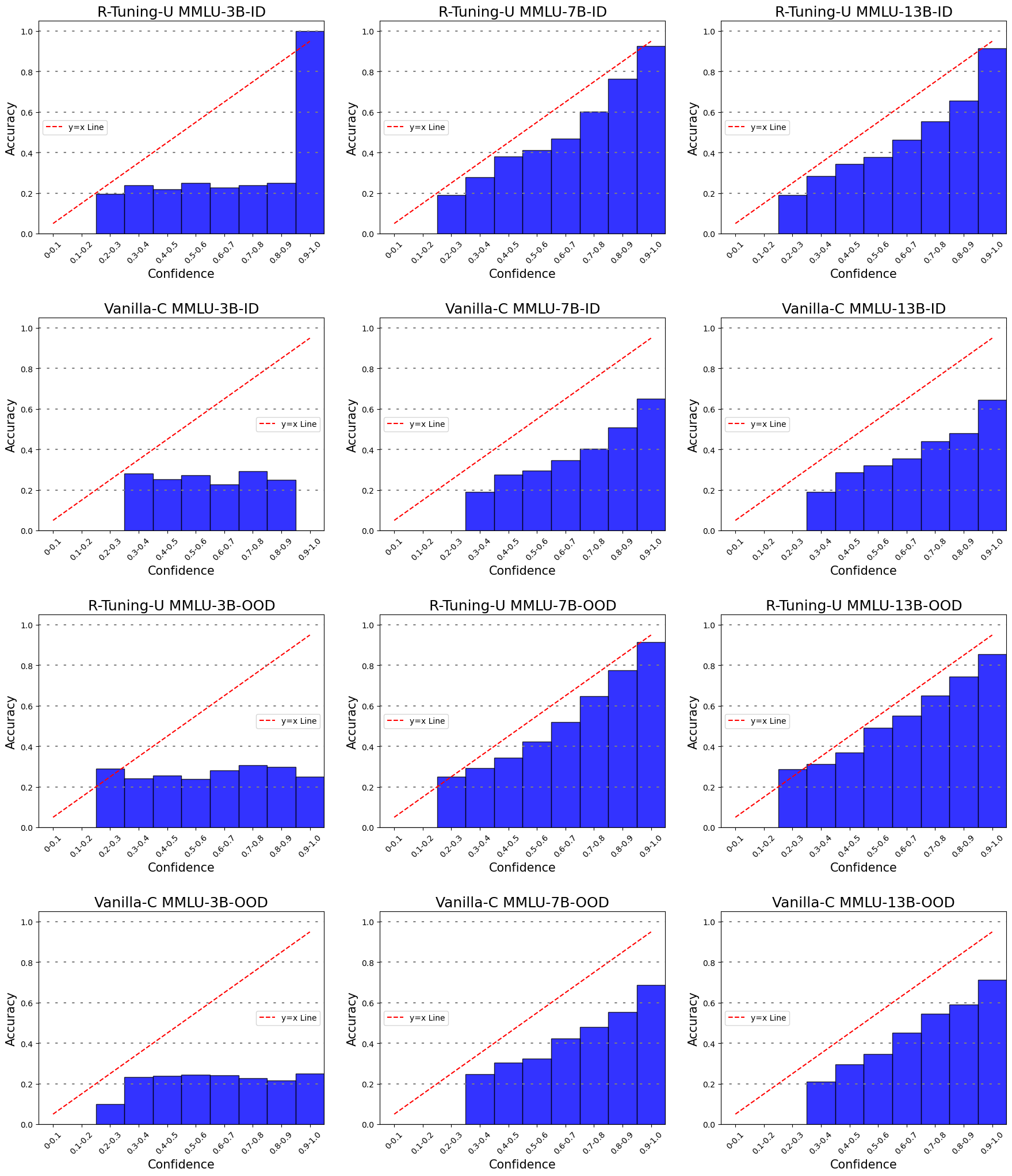}
    \caption{The ECE (Expected Calibration Error) on MMLU dataset of {\ModelUncertain} and {\vanillaC}.}
    \label{mmlu_ece}
\end{figure*}

\begin{figure*}[t]
    \centering
    \includegraphics[width=\textwidth]{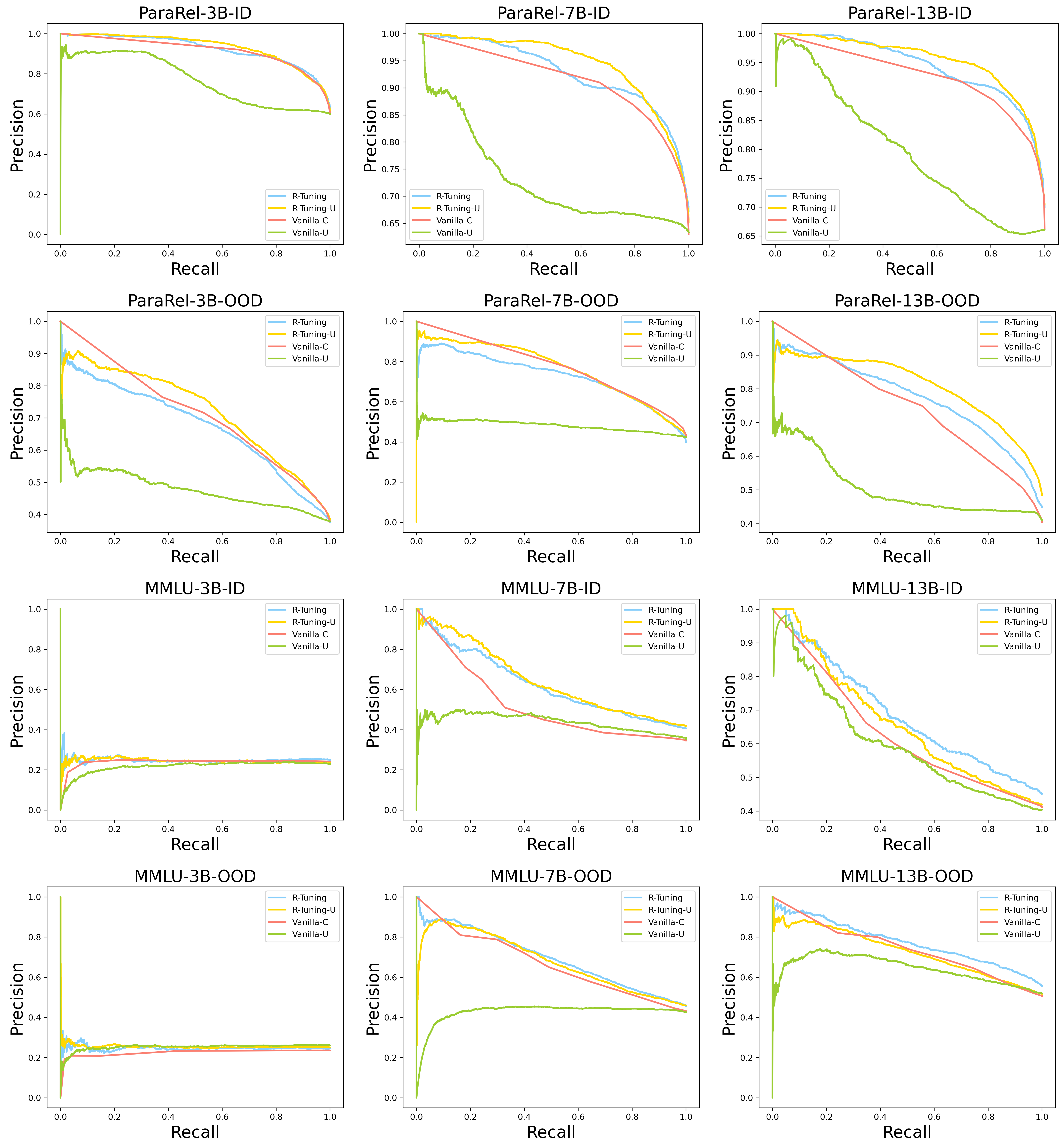}
    \caption{The AP curves of {\ModelName}, {\ModelUncertain}, {\vanillaC}, and Vanilla-U on ParaRel and MMLU datasets.}
    \label{fig:ap_curves_4_types}
\end{figure*}

\end{document}